\documentclass{article}
\usepackage[
  letterpaper,
  margin=1.5in
]{geometry}
\usepackage[utf8]{inputenc}
\usepackage{graphicx} % Required for inserting images
\usepackage[T1]{fontenc}   % recommended for yfonts
\usepackage{yfonts}        % provides \textswab, \textfrak, \textgoth
\usepackage{mathrsfs}      % provides \mathscr
\usepackage[numbers,sort&compress]{natbib} % or omit 'numbers' for author-year
\usepackage{amsfonts}
\usepackage{amsmath}
\usepackage{amssymb}
\usepackage[dvipsnames]{xcolor}
\usepackage{authblk}
\usepackage{comment}
\usepackage{pdflscape}

\usepackage{booktabs}
\usepackage{array}
\usepackage{xcolor}
\usepackage{colortbl}
\usepackage{enumitem}
\usepackage{hyperref}
\usepackage{pifont}
\usepackage{makecell}
\usepackage{titlesec}
\usepackage{multirow}

\usepackage{soul}
\usepackage{xurl}          % smarter URL line breaks
\usepackage{proof}
\usepackage{mathtools}
\usepackage{tikz}
\usepackage{tcolorbox}
\usepackage{float}

\usepackage[table,dvipsnames]{xcolor}
\usepackage{array}
\usepackage{booktabs}
\usepackage{tabularx}
\usepackage{ragged2e}

\usepackage{caption}
\renewcommand{\arraystretch}{1.2}

\usetikzlibrary{positioning, calc, shapes, arrows.meta, fit, backgrounds, decorations.pathmorphing}
\usepackage{amsthm}
\theoremstyle{definition} % Makes the text upright (not italic)
\definecolor{HE_Form}{RGB}{0, 0, 200}       % Blue for Formation
\definecolor{HE_Intro}{RGB}{0, 150, 0}      % Green for Introduction
\definecolor{HE_Elim}{RGB}{230, 100, 0}     % Orange for Elimination
\definecolor{HE_Comp}{RGB}{150, 0, 150}     % Purple for Computation

\def\boldclass{\bf\sf}
\def\P{{\boldclass P}}
\def\NP{{\boldclass NP}}

\newcommand{\G}{\mathcal{G}} % a General hypergraph of mathematics
\newcommand{\HE}{\mathcal{E}} % hyperedges
\newcommand{\uproof}{\mathcal{P}} % the Universal hypergraph
\newcommand{\U}{\mathcal{U}} % the Universal hypergraph
\newcommand{\UU}{\mathscr{U}} % the collection of universal hypergraphs
\newcommand{\C}{\mathcal{C}} % Contingent mathematics
\newcommand{\St}{\mathcal{S}} % Statements
\newcommand{\T}{\mathcal{T}} % Tools for reasoning, i.e. lemmas which can be freely applied.
\newcommand{\N}{\mathcal{N}} % neighborhood
\newcommand{\Hum}{\mathcal{H}} % Human mathematics as a subgraph of U
\newcommand{\HH}{\mathscr{H}}
\newcommand{\CO}{\mathcal{O}} % standard math notation
\newcommand{\CT}{\mathcal{T}} % application of a rule or hyperedge

\newcommand{\Proof}{\bf Pr} % the proof map

\usepackage{listings}
\usepackage{xcolor}

\definecolor{codegreen}{rgb}{0,0.6,0}
\definecolor{codegray}{rgb}{0.5,0.5,0.5}
\definecolor{codepurple}{rgb}{0.58,0,0.82}
\definecolor{backcolour}{rgb}{0.96,0.96,0.96}

\lstdefinestyle{pseudocode}{
    backgroundcolor=\color{backcolour},   
    commentstyle=\color{codegreen}\itshape,
    keywordstyle=\color{magenta}\bfseries,
    numberstyle=\tiny\color{codegray},
    stringstyle=\color{codepurple},
    basicstyle=\ttfamily\scriptsize, % Use monospaced font
    breakatwhitespace=false,         
    breaklines=true,                 
    captionpos=b,                    
    keepspaces=true,                 
    numbers=left,                    
    numbersep=5pt,                  
    showspaces=false,                
    showstringspaces=false,
    showtabs=false,                  
    tabsize=4,
    frame=lines,                     % Adds top and bottom lines
    language=Python,                 % Syntax highlighting for Python
    mathescape=true,                  % CRITICAL: Allows $\mathcal{U}$ in code
    xleftmargin=-1cm,   % Pulls the left edge 1cm into the margin
    xrightmargin=-1cm,  % Pulls the right edge 1cm into the margin
    resetmargins=true   % Ensures these apply correctly}
}

\newcommand{\maissam}[1]{{\color{red}\footnotesize{MB: #1}}}

\definecolor{TitleBlue}{HTML}{143B5F}
\definecolor{SoftGray}{HTML}{5A6673}
\definecolor{LightRule}{HTML}{D8E1EA}
\definecolor{TableHead}{HTML}{EEF3F7}
\definecolor{Good}{HTML}{2F7D32}
\definecolor{Warn}{HTML}{B26A00}
\definecolor{Bad}{HTML}{A12A2A}

\definecolor{darkblue}{RGB}{38,64,115}
\definecolor{medblue}{RGB}{55,80,130}
\definecolor{lighterblue}{RGB}{235,241,250}
\definecolor{mygreen}{RGB}{26,154,26}
\definecolor{myred}{RGB}{179,26,26}
\definecolor{myamber}{RGB}{204,136,0}

\newcommand{\yes}{\textcolor{mygreen}{\ding{51}}}
\newcommand{\no}{\textcolor{myred}{\ding{55}}}
\newcommand{\pmark}{\textcolor{myamber}{\raisebox{0.1ex}{\textbf{\texttildelow}}}}

\newcommand{\tri}[3]{#1\;#2\;#3}

\newcolumntype{Y}{>{\RaggedRight\arraybackslash}X}
\newcolumntype{C}[1]{>{\centering\arraybackslash}p{#1}}

\title{Artificial Intelligence and the \\ Structure of Mathematics}

\author[1,2]{Maissam Barkeshli,}
\author[3]{Michael R. Douglas,}
\author[3,4]{\authorcr Michael H. Freedman}

\affil[1]{Meta Superintelligence Labs, Fundamental AI Research }
\affil[2]{Department of Physics and Joint Quantum Institute, \authorcr University of Maryland, College Park}
\affil[3]{Center for Mathematical Sciences and Applications, \authorcr Harvard University}
\affil[4]{Logical Intelligence}

\date{}

\begin{document}

\maketitle

\begin{abstract}
Recent progress in artificial intelligence (AI) is unlocking transformative capabilities for mathematics. There is great hope that AI will help solve major open problems and autonomously discover new mathematical concepts. In this essay, we further consider how AI may open a grand perspective on mathematics by forging a new route, complementary to mathematical\textbf{ logic,} to understanding the global structure of  formal \textbf{proof}\textbf{s}. We begin by providing a sketch of the formal structure of mathematics in terms of universal proof and structural hypergraphs and discuss questions this raises about the foundational structure of mathematics. We then outline the main ingredients and provide a set of criteria to be satisfied for AI models capable of automated mathematical discovery. As we send AI agents to traverse Platonic mathematical worlds, we expect they will teach us about the nature of mathematics: both as a whole, and the small ribbons conducive to human understanding. Perhaps they will shed light on the old question: "Is mathematics discovered or invented?" Can we grok the terrain of  these \textbf{Platonic worlds}?
\end{abstract}

\tableofcontents
\bigskip

\section{Introduction}

Artificial intelligence (AI) has demonstrated revolutionary progress in mathematical reasoning. Multiple different neural reasoning models \cite{Metz2025-nyt-imo-gold} -- Google DeepMind's Gemini \cite{DeepMind2025-IMO-Gold} and OpenAI's undisclosed GPT model -- announced gold medal performance on the 2025 International Math Olympiad (IMO). Just a few months later, several AI systems essentially saturated Putnam-level problem sets \cite{AxiomMathPutnam2025,liu2026numinaleanagentopengeneralagentic,PutnamBenchLeaderboard,Tsoukalas2024PutnamBench}. The most difficult math benchmark to date, FrontierMath \cite{GlazerEtAl2024FrontierMath}, created by experts to maximally challenge modern AI systems at mathematical reasoning, has seen many of its highest tier problems solved by AI \cite{epochai_frontiermath}. AI agents have also been indispensable in completing formal proofs of the prime number theorem \cite{mathinc_gauss_2025} and higher dimensional sphere packing \cite{mathinc_spherepacking_lean}, contributing hundreds of thousands of lines of Lean code and dramatically advancing Hilbert's dream of formalizing mathematics. 
FirstProof \cite{abouzaid2026first}, a set of 10 novel research-grade problems that were held-out intermediate results in the work of expert mathematicians saw the majority of its problems solved autonomously \cite{feng2026aletheiatacklesfirstproofautonomously, OpenAI2026FirstProofSubmissions} within a week of release. Researchers in mathematics, computer science, and theoretical physics have found modern AI systems useful in the wild, for developing proofs and providing high level reasoning to aid their work \cite{vanraamsdonk2025finiteentropysumsquantum,Tao2025-mathstodon,ghrist2024latticevaluedbottleneckduality,jang2025pointconvergencenesterovsaccelerated,schwartz2026,bryan2026motivicclassspacegenus,BloomErdosProblems,guevara2026singleminusgluontreeamplitudes,brenner2026solvingopenproblemtheoretical,coester2026transpositionnearlyoptimaliid}.

We stand at a step change in mathematical research \cite{FraserEtAl2024WillMachines,DeDeo2024AlephZero,Davies2024WorkingWithMachines,DeToffoli2024ProofsForAPrice,McLarty2024PoincareReasoning,Ochigame2024AutomatedMath,PoggioFraser2024CompositionalSparsity,BengioMalkin2024AItheory,venkatesh2024some,naskrkecki2025mathematical,yang2026formal,Avigad2026MathematiciansAgeAI}, with no clear limits on the autonomy and creativity of our AI agents. Can AIs go beyond lemma building - to settle human-posed problems (something we now have learned to expect!), to autonomously discover important new mathematical structures on their own? How would AIs traverse the mathematical world and know what is worth reporting back to us? Imagine by analogy that we send an unmanned vehicle to a new planet. The first thing we might want to know, particularly for designing the vehicle in the first place, is what the terrain of this planet is like. There are high level considerations like, "Is the planet round?" and also details like, "There is a lake of He-3 at the south pole." which we would regret not learning. So then, what do we already know about the terrain of the Platonic worlds, and what might we ask our AI agents to look for when they arrive? Can we abstract for the AIs what is special about our small ribbon of "human mathematics" (HM), within the universe of all possible deductions?

This essay attempts to look ahead to the design of autonomous mathematical discovery agents -- our AI ``space-probe" -- and grapple with several basic questions that arise. First, we will sketch a formal representation of mathematics in a graph theoretic language, along with conceptual tools to work with it. These tools are taken from logic, computer science, mathematics, machine learning and AI. We hope to  formalize questions such as: What determines the small part of the vast structure of provable mathematics which actually corresponds to human mathematics (existing and potential)? Can we define AI agents which could (re)discover this portion and not get lost? What role is played by considerations beyond proof: definitions, conjectures, judgments of interestingness, taste, and so on? Clearly these are deep and hard to answer questions; while we will speculate about the answers, our main goal is simply to convince the reader that they can be studied in a precise way. 

\subsection{Outline}

We begin the main discussion in \S \ref{s:hyper}, where we define hypergraphs which represent the compositional and deductive nature of mathematics. Hyperedges in these graphs represent deductions and the construction of statements, functions, data types and proofs from elementary constituents. One could choose from many foundations on which to base this; we follow dependent type theory as used in the Coq/Rcoq and Lean theorem proving languages, in the version given in Chapter 1 of \cite{aczel2013homotopy}. Using this framework, we define several hypergraphs.  The most fundamental is the {\bf universal} hypergraph $\U$ of all provable statements in a given foundation.  We also discuss the structural hypergraph $\mathcal{S}$ of more general mathematical objects outside of $\U$ and we provide a discussion of abstractions and complexity measures in this framework. $\U$ is infinite but can be thought of as a colimit of locally finite sub-hypergraphs, starting from the axioms and building up all conceivable mathematics. Its structure as a mathematical object is one answer to the question ``what is the structure of mathematics?'' However it is computationally intractable: it has (at least) doubly exponential growth, and even finding the statements of human mathematics within it is challenging. Still, one could formulate our first question as: {\it are there natural mathematical (not necessarily tractable) conditions, formulated purely in terms of the hypergraph $\U$ and not in terms of the meanings we attribute to it, which constrain the sub-hypergraph $\Hum\subset\U$ of human mathematics? } This and other questions about the universal nature of mathematics are discussed in \S \ref{s:universal}.

It would be a bold claim indeed that mathematics is so constrained that the nature of human mathematics and the course of its discovery are predetermined, and we will not assume this. More plausibly, the situation is analogous to that of evolutionary biology.  There, while natural law and the workings of natural selection are powerfully constraining, one can make a strong case that many features
of the history of life are contingent, depending on chance events which could have gone otherwise \cite{gould1989wonderful}. Thus our framework must also be able to describe the {\bf contingent} aspects of mathematics and its development. We will model the history of mathematics with a series of hypergraphs $\C_t$ for $t=0,1,\ldots$ representing the statements topical at time $t$, including proven statements, conjectures, data types and concepts. Here $t$ is not necessarily historical time, but that is an important special case.

Now, we know that humans can discover mathematics, and we believe that AI's will do so very soon. Both satisfy physical constraints and thus our models of how mathematics is discovered must include computational constraints. We enforce the computational constraints by requiring $\C_t$ to grow at a bounded rate, and that the processes which govern its time evolution be computationally tractable. We discuss models of autonomous mathematical discovery (AMD) in \S \ref{s:models}.  Many models and systems use the framework of reinforcement learning, which we briefly review. We then discuss issues specific to mathematics and AMD, such as conjecturing, proof and abstraction.

Systems for discovering mathematics are a classic topic in AI, and early famous works include AM \cite{Lenat1977IJCAI}, HR \cite{colton2012automated} and Graffiti \cite{fajtlowicz1988conjectures}. Some recent systems which inspired us are 
Minimo \cite{poesia_learning_2024}, Dreamcoder \cite{ellis_dreamcoder_2020},
Lilo/Stitch \cite{grand_lilo_2023,bowers_top-down_2023} and Fermat \cite{tsoukalas_learning_2025}. We survey these systems, abstracting away many of the (important) details of real AMD systems to draw a simple but hopefully useful picture.

In \S \ref{s:human}, we try to fit these pieces together and speculate about human mathematics. Set against the combinatorial explosion of mathematics, the fundamental computational constraints on AIs suggests that humans will always have untouched directions left to explore, and will do so parsimoniously with AIs.

In \S \ref{s:paths} we outline some paths forward. We provide some thoughts on the coming of a new discipline, which one might call computational metamathematics, enabled by advances in AI and formalization programs. In \S \ref{ss:criteria} we discuss criteria by which to judge automated mathematical discovery, and offer our list in Fig. \ref{fig:criteria} to help assess the capabilities of advanced AI for math systems in the coming years. In Table \ref{table:amd_ratings}, we have various frontier models rate a number of AMD systems according to our criteria, which may be useful for measuring recent progress.

\def\amdsystem{agent}
\def\anAmdsystem{an Agent}

\begin{figure}[H]
%    \centering
    \begin{tcolorbox}[title=Criteria for \anAmdsystem\ for Autonomous Mathematical Discovery] % Title is optional
\begin{enumerate}
    \item The \amdsystem\ must work with an open-ended mathematical language, able to express new theorems, proofs and concepts.
    It need not go beyond human knowledge
    to make discoveries, just beyond its starting knowledge, but its design should make this possible in principle.
    \item The \amdsystem\ produces verifiable formal or informal proofs.
    \item Given a theorem, the \amdsystem\
    can tell whether it is new, meaning not easily provable from its current knowledge. 
    In the terms of \S \ref{ss:efficiency}), this means large proof complexity $m(\bar{P})$ relative to the current corpus $\mathcal{C}_t$.
    \item The \amdsystem\ proposes new theorems and proves them.
    \item The \amdsystem\ proposes new definitions of functions, data types and concepts.
    \item The \amdsystem\ selects a small number of its proposals as particularly interesting -- these are its discoveries.
    \item The \amdsystem\ gives reasons for its selection.  These could be stated in natural language, but need not be as
    long as they contribute to achieving other criteria.  
    For example, the \amdsystem\ might estimate the value of each new theorem by quantifying how much it compresses the proof hypergraph--optimizing the efficiency measure $E(\bar{P})$ or the abstraction utility $U(A)$ (\S \ref{ss:models})--and use this to rank its proposals.
    \item Closely related to (7): The \amdsystem\ should be able to produce not just a proof, but a research program. This requires both sensing what statements may be in reach,  and the taste to evaluate them. 
    \item There should be independent criteria validating the choice of discoveries.
    These include human evaluation and tests of the sort discussed in \S \ref{ss:tests}.
%\global\edef\Nvalid{\theenumi}
    \item Suppose the \amdsystem\ is run in a closed loop, generating new theorems and concepts, adding its discoveries to its knowledge base, and
    repeating the process.  Each iteration should produce new discoveries, leading to an unlimited expansion of its knowledge.
\end{enumerate}
    \end{tcolorbox}
    \caption{Criteria for AMD}
    \label{fig:criteria}
\end{figure}

\section{The hypergraphs of mathematics}
\label{s:hyper}

Our goal in this section is to briefly sketch the overarching structure of formal mathematics in terms of a series of hypergraphs. We consider it to be an important research program to further flesh out and make rigorous the discussion below. We subsequently use the hypergraphs to pose fundamental questions about the structure of mathematics and consider how AIs and humans traverse these worlds. 

\subsection{Universal proof hypergraph}

A formal mathematical system determines the structure of a directed, ordered, colored, acyclic hypergraph (henceforth ``hypergraph'').\footnote{
The idea of representing logical arguments graphically originated with C. S. Pierce. There are many ways to do it; a few include \cite{Girard1987,Dechter2003,GeuversLoeb2007}. 
} We review this now, focusing on a universal hypergraph $\U$ which contains all provable statements. Technical differences between hypergraphs and graphs present an obstacle to turning mathematics into a metric space and then thinking geometrically; hypergraphs involve n-ary relations whereas geometry, generalizing "graph",  is founded on a binary operation, distance.\footnote{The richest expression of geometry is Riemannian geometry (or possibly its Finsler generalization). As we discuss later, could these subjects have n-ary analogs to which appropriate sequences of hypergraphs could limit?}
$\U$ is infinite and we will show that it generally has doubly exponential growth.

Start with a formal \textbf{symbolic language}, a finite \textbf{set of axioms}, and \textbf{deductive rules.} 
Let $\bar{\St}$ be the set of all expressions in the symbolic language.
We will say more about this in \S \ref{ss:structure},
but a concrete example to keep in mind is first order logic and Peano arithmetic (PA).
We recall that in PA, the natural numbers are defined inductively as either $0$ or the successor $Sx$ of a natural number $x$. Expressions which can be true or false are called propositions; examples in PA are $0=0$, $0=S0$, $S0+S0=SS0$, $(\exists x, x=S0)$ and so on.
Expressions also include values like $0$ or $S0$, and expressions like $S0+S0$, but for now we consider the propositions.

The vertices (nodes) of the universal hypergraph $\U$ are the subset of all \textbf{provable} propositions in $\St$. A directed hyperedge of type $(p,q)$ (sometimes called a directed hyperarc) within the hypergraph is a list of vertices, divided into $p$ ``input'' vertices and $q$ ``output'' vertices, the deductive consequences of the input. A hyperedge captures how some propositions are combined using a deductive rule to derive new propositions, and carries a color denoting a particular rule. For example, if $A$ and $B$ are two nodes in the graph, then we have a directed hyperedge \textbf{$(A, B \vdash A \wedge B)$}. Another example is modus ponens:  if two previous vertices  are $A$ and $A \Rightarrow B$, respectively,  then we can add a hyperedge with these inputs and outputting $B$. Note that $A$ and $A\Rightarrow B$ enter differently, so we need to remember the order of the input vertices: this is an ordered (or labeled) hypergraph. We allow the rules to look at the propositions and only apply if conditions are met; also they can come with options which specify its operation. For example, a rule ``equals can be substituted for equals'' could take as inputs $A=B$ and an expression $s$ containing $A$, and produce an output in which $A$ is replaced with $B$. This rule can only be applied if $s$ contains $A$, and might come with an option specifying which of multiple occurrences of $A$ to replace. However, it is essential that all rules have only finitely many inputs and can be applied in only finitely many ways.\footnote{There are many variations on this construction.  For example, by rearranging the deductive rules and introducing new vertex types, one could reduce each inference rule to a concatenation of hyperedges of types $(1,1)$ and $(2,1)$. In Boolean logic, for example, NAND (Sheffer stroke) and NOR (Peirce arrow) are each universal. One can even avoid hypergraphs altogether, and instead use a directed bipartite graph. To do this one adds a new class of nodes (call them "red"), one for each hyperedge and then replace each  $(p,q)$ hyperedge by  $p$ arrows into the corresponding red node and $q$ arrows out of that red node (we leave the generalization to ordered hypergraphs as an exercise).  By various gadgets one may inject hypergraphs into even simpler structures, such as ordinary unoriented graphs, but there seems little to be gained by such encodings. After all, anything can be encoded in a a string of zeros and ones.}

Axioms (denoted $\U_0$) correspond to root nodes in the hypergraph, which we picture as lying at the bottom of the hypergraph. A theorem is an implication node of the form (hypotheses $\Rightarrow$ conclusions).
Every non-root node $X\in \U$ has at least one proof, a backward closed hypergraph originating on the root nodes and terminating on $X$. One can also talk about conditional proofs which also depend on non-root nodes, the hypotheses. Suppose we have a hypergraph proving $X$ given the hypothesis $A$, denoted $A \vdash X$.
Then there is a deductive rule (the rule of introduction for $\Rightarrow$) which creates the corresponding implication node $A \Rightarrow X$.\footnote{ An operation which takes an entire sub-hypergraph as input could not be considered to be a hyperedge. Rather, we take the input and output nodes of the sub-hypergraph as its inputs.  Thus it is a hyperedge but one which depends in a nonlocal way on its inputs.
}
Note that many proofs are hypergraphs which are not 
sub-hypergraphs of $\U$, only of a larger structural hypergraph that we define in the subsequent section $\St$. Namely, those for which the hypotheses are not universally provable. One such class of hypotheses are open formulas such as ``$x$ is prime'' which we return to below. But mathematics also includes provable theorems whose hypotheses are propositions which are not known to be provable; consider a theorem proven conditionally on the Riemann Hypothesis. These are hints that to fully discuss mathematics we need to go beyond $\U$, and more reasons will appear below. Still, $\U$ is universal and has a special role in the discussion.

There is a conceptual simplification possible if we are only discussing a single proof. Then the more familiar concept of a directed graph is adequate. Each vertex is a necessary intermediate proposition $s$, and the totality of (say $p$) arrows pointing to $s$ collectively represent a $(p,1)$ hyperedge in the HG notation. This works fine for a single proof with no redundant internal implications, but  to diagram all (or even several) proofs at once,  the grouping of the arrows needs to be remembered and this is the essence of the hyperedge formalism. Notably, MathLib, the largest Lean4-based mathematical library, is structured as a directed graph.

Starting with the axioms, we can consider a sequence of hypergraphs $\U_t$, where $t$ is the depth of the hypergraph, which is the minimal number of deductions required to reach every node in $\U_t$. We can ensure $\U_t$ is finite, although this requires some care. One complication is variable specification -- replacing a variable $x$ with a particular natural number. In this case, there are a priori an infinite number of specifications, which would make $\U_t$ locally infinite. We can restore local finiteness by only allowing a finite number specifications at each layer of deductions, starting with the axioms. This can be aided by the structural hypergraph, discussed below. Our universal hypergraph $\U$ is then the colimit of $\U_t$ as $t \rightarrow \infty$. 

The doubly exponential nature of growth with depth can then be seen by considering the silliest example within propositional calculus. Start with propositions $A_1, ...,A_k$ and at each layer just take all pairwise AND of propositions built (just) at the immediately preceding layer.  Ignoring small combinatorial factors the number of propositions grows like: $k, k^2, (k^2)^2=k^4, (k^4)^2=k^8,..., k^{(2^j )}$ after j steps. We see that the length of propositions are potentially growing exponentially and the total number doubly exponentially (in step number).\footnote{ The growth as a function of the number of nodes $s$ of the graph is $k^{s^2}$ (\cite{wigderson2019mathematics} \S 5.2.1).}
This silly example also illustrates how little we want to actually hear about when our AI space probe reports back. 

\subsection{Structural hypergraphs}
\label{ss:structure}

While one can work with a universal hypergraph $\U$ using just the definitions so far, 
there are many ways in which it can be improved.  One observation is that while we are representing logical structure in the hypergraph, other structures -- of equations, definitions, concepts such as that of an abstract group or ring, and anything else -- are all left for the symbolic language to describe.

If we go this route, then the structure of $\U$ will depend very much on the details of the symbolic language: which rules of deduction can be applied to a given subgraph, and which output nodes they produce. This becomes very clear once one goes to first order logic, with functions, predicates, values and variables.
Here one needs rules of deduction which define and apply functions,
specialize and make substitutions for variables, and so on.  Consider ``equals can be substituted for equals,''
which could correspond to a $(2,1)$-hyperedge with (say) inputs $A=B$ and a second statement which contains $A$.
While the substitution $A\rightarrow B$ looks easy to state at first, there are subtleties: say the second statement
is $\forall x, A=C$ where $C$ refers to $x$, while $B$ in the first statement $A=B$ uses $x$ to mean something else.
Then there is a danger of misinterpretation (variable capture).
Such subtleties can be resolved, but to do this we need to give many details about the symbolic language.

To deal with such issues, modern logical frameworks (including those used in ITP systems) systematically build up the expressions in terms of structural rules. For example, $A=B$ is composed of the expressions $A$ and $B$; this construction could be associated to a hypergraph with nodes representing $A$, $B$, and $A=B$ connected by an ``$=$'' $(2,1)$-hyperedge.  By representing all of the compositional operations this way, one can represent all of the structure of the expressions graphically. In this way, the set of all statements $\bar{\St}$ becomes the set of nodes of what we will call the {\bf structural} hypergraph $\St$, and the finite set of hyperedges become rules of construction that iteratively create all possible statements in $\bar{\St}$ from an initial set $\bar{\St}_0$. While issues such as variable capture are still present, one can use the hypergraph to help resolve them. For example, a hyperedge would refer not to the string ``$x$'' but to the node corresponding to $x$.

Within $\St$, the propositions are distinguished as statements which can be true or false.  Let's denote them as $\Pi\St$, and more generally $\Pi\G$ is the sub-hypergraph of $\G$ which keeps all propositions and hyperedges between them.

In a full development of this type, textual representations of the statements are redundant; the expression $\bar s$ corresponding to the node $s$ is entirely determined by the sub-hypergraph with terminal node $s$. This means that a given node $s$ determines a series of equivalent expressions, corresponding to the sub-hypergraphs that end on $s$. This is also closer to the way statements are represented in ITP systems.  We recall that the first step in processing a high level language (programming, theorem proving, or others) is to parse the input statements and produce a parse tree explicitly representing this compositional structure.  The resulting internal structure is usually not a pure hypergraph, and makes use of other data structures such as symbol tables.  But it could be formulated as a hypergraph, and such a formulation of the internal structure produced by an ITP system is what we have in mind. 

\subsection{Abstraction}
\label{ss:abstraction}

The business of mathematics is {\bf abstraction}, the ability to take a complicated construction
and treat it as a single object of thought.  This shows up in many ways.  We discussed one way earlier: a
theorem can have two representations: a sub-hypergraph with inputs the hypotheses (say $A$ and $B$) and outputs the conclusion (say $X$), or a single implication node\footnote{
Conventionally, $\Rightarrow$ associates to the right, so this is $A \Rightarrow (B \Rightarrow X)$.}
$A \Rightarrow B \Rightarrow X$, a shorter and ``more abstract'' representation of the result.
Given the implication node, one can reprove the same result in another context
with a very small hypergraph.  Furthermore one can use the same implication node many times.  For a ``real world'' agent carrying out
the arguments, this could clearly give huge savings.  The abstract representation also hides the details of the proof, which can have advantages (and disadvantages) for reasoning.  Conceptually, it allows ``coarse graining'' from a fully detailed ``microscopic'' definition of mathematics to a ``macroscopic'' definition.

Another form of abstraction is function definition.  Given a complex formula $f$ written in terms of inputs $x$ and $y$, one can abstract it into a definition of a function $x, y \rightarrow f(x,y)$.  This is formally  parallel to the abstraction which creates a theorem -- a formula is a sub-hypergraph of the structural hypergraph, and we are abstracting it as a single ``function'' node.\footnote{ Once we can define functions, we face the problems of non-termination and uncomputability. In our specific framework (dependent type theory), we avoid this problem by enforcing strong normalization: all functions which follow the typing rules and thus are defined in $\mathcal{S}$ will terminate.  Without going into details, this is because
only certain types of infinite definitions are permitted, such as the recursion operator
used in appendix \ref{s:exhyper}.  Potentially infinite processes like an infinite loop or the Collatz process cannot be represented, though one can approximate them by introducing an additional ``fuel parameter'' which counts the number of steps and terminates when some predetermined limit is reached.  This ensures that every node in $\mathcal{S}$ has a finite construction history.}
And in some logical formalisms, notably those using the Curry-Howard correspondence, the two types of abstraction are the same: a theorem is a function which accepts the proofs of its hypotheses as inputs, and outputs a proof of the conclusion.

Data types can also be abstractions.  The classic example used to illustrate the value of dependent type theory is that of a vector, defined as an $n$-tuple of elements of a field $\Bbbk$ or simply $\Bbbk^n$.  Intuitively the vector type is a function of two arguments, $n$ and $\Bbbk$, but making this work in a fully consistent way is surprisingly difficult.\footnote{Readers familiar with C++ can compare with its concept of template.  This allows defining a vector type which can be used to declare variables and functions, but such a type cannot be assigned to a variable or passed as an argument to a function (types are not first class).} In the Lean and Rocq ITPs based on dependent type theory, the vector type is a function of $n$ and $\Bbbk$, defined like any other. And, the notion of data type is further extended to associate functions, predicates and axioms with specific types (the structure or class concept).  One can for example define an abstract group as a class with a carrier type (the elements), functions (the multiplication and inverse laws) and the group axioms.\footnote{
In set theoretic foundations this would be a signature-axiom class, as in Bourbaki.}

Let us explain how we represent an abstraction. As we mentioned earlier, given a sub-hypergraph $A,B,\ldots\vdash X$ there is a deduction rule producing the node $A \Rightarrow B \Rightarrow \ldots \Rightarrow X$.\footnote{One can reduce this to a single rule in terms of $A \vdash B \Rightarrow \ldots \Rightarrow X$.} This can be used with modus ponens (perhaps applied several times) in place of the original proof sub-hypergraph.\footnote{An alternate representation would be to introduce a new hyperedge with the same inputs and outputs as the sub-hypergraph.  This is largely equivalent to the representation using nodes, but this point of view helps one see why different choices of rules of construction and deduction can lead to equivalent hypergraphs.}  
Similarly, a function sub-hypergraph $x,y\rightarrow f(x,y)$ can be turned into a function node $f$ using the ``lambda abstraction,'' and applied using a function application $(2,1)$-hyperedge twice.\footnote{The first application turns $f$ and $x$ into $f(x,\cdot)$, and the second takes
this and $y$ and produces $f(x,y)$.}  By using the new node representing the abstraction, one can make a new, shorter proof or construction of a result. Doing this has advantages and disadvantages -- while proofs become shorter, the branching factor (number of ways
to apply the rules) will become larger and the rate of growth of the hypergraph with depth will go way up.  Thus the choice of which abstractions to make is very important, as we discuss below.

\subsection{Proof objects and the hypergraph of proofs}
\label{ss:proofobj}

Our discussion so far made a shift of perspective.  We started with a universal hypergraph $\U$ of proven propositions. We then changed focus to a larger universal hypergraph $\St$ of all expressions, containing all propositions $\Pi\St$ as well as other abstractions, with $\U\subset \Pi\St$ as a sub-hypergraph.

Now we will make another, final shift of perspective.  In addition to propositions, values, and functions, we will represent {\bf proofs} as nodes in a hypergraph. Let us start out by simply granting this, and return below to how it can be done.  Thus, we introduce a new subtype of expression called ``proof.''  Proofs are denoted $\bar p$, and their associated nodes as $p\in\St$.  Each proof is the proof of a proposition $\bar P$ with node $P$. This relation is denoted $p:P$ and represented by a $(1,1)$-hyperedge with input $p$ and output $P$.  The inverse
relation is not a hyperedge (there is no direct way to construct $p$ from $P$), but there is a way to test whether a given $P$ has any associated proof node or not.

In this framework, we still have $\U\subset \Pi\St$, but now $\U$ can be inferred from the proof nodes and the test we just discussed.  This picture is more natural in several ways.  It explicitly represents the subset of proven nodes as the image of a map, instead of implicitly through the relation between two hypergraphs.  It naturally allows for multiple proofs of the same statement, so one can discuss questions such as ``when are two proofs to be considered the same.''\footnote{This is the starting point for homotopy type theory and related developments.  Most ITP systems (including Lean 4) do not use this idea, and only distinguish proven and unproven propositions.}

Importantly, the rules of deduction can be a special case of rules of construction which act on proofs and produce proofs.  One example is the rule $A, B \vdash A \wedge B$.  In terms of proof objects, a proof of $A \wedge B$ is simply a pair of proofs, $a$ of $A$ and $b$ of $B$.  Thus, it can be constructed using the pair constructor, which takes two general statements $a,b$ and produces the pair $(a,b)$.  

Elaborating on this idea, all of the rules of deduction can be interpreted as rules of construction. This picture arises naturally in type theory, where propositions are types and proofs inhabit these types. This is the Curry-Howard correspondence.  We need not go into details on how to use the proof object concept, but it is a way to precisely define the proof nodes.

The upshot is, the graph $\U$ of our earlier discussion is the ``shadow'' of
a universal hypergraph of proofs $\uproof\subset\St$.
Its root nodes are the axioms $\uproof_0$, to which hyperedges can be added to construct new proof nodes.
Each new proof node $p$ has a type $p : P$ (``$p$ inhabits the type $P$'') and each new $P$ is a new proposition in $\U$. 

\subsection{Equality, isomorphism and canonicalization}
\label{ss:equality}

A next step in the coarse graining is to represent equivalence relations and quotients. An equivalence relation can be represented mathematically as a map $R: A \rightarrow (A \rightarrow \mathcal{U})$, such that $R$ is reflexive, symmetric, and transitive. R maps a type $A$ to proofs that certain of its inhabitants are "equivalent." In the simplest case we can think of this as defining a proposition for each $R(a)(b)$, where $a, b$ are each of type $A$, whose truth value determines whether $a$ and $b$ are equivalent. The more general map into $\mathcal{U}$ can encode additional structure, such as the proof of how $a$ and $b$ are related. For example given $a = 3$ and $b = 2+1$, $R$ determines the proof that they are equivalent. 

Given a type $A$ and an equivalence relation $R$, we introduce a new node in $\St$ representing the quotient type $Q = A/R$. This node serves as the generic container for the equivalence classes. For every term node $a : A$, there exists a \textbf{Projection Hyperedge} (denoted $\texttt{class}$ or $[\cdot]$) connecting $a$ to a term node in $Q$.
\[ a \xrightarrow{\texttt{class}} [a] \]
If we have a sub-hypergraph that only depends on $a$ up to equivalence, then we can obtain a quotient subgraph that only depends on the node $[a]$. If $a$ and $b$ are equivalent ($a \sim b$), then $[a]$ and $[b]$ are by definition the same node in the quotient sub-hypergraph.

The new node $[a]$ is opaque---an ``abstract'' value. To compute with it (i.e., to define a function $f$ out of $Q$), we cannot inspect the node directly. We must define the function on the original nodes ($f' : A \to B$) and provide a \textbf{Consistency Hyperedge} (Proof of Congruence) showing that equivalent inputs yield equal outputs:
\[ \forall x, y.\; R(x,y) \to f'(x) = f'(y) \]
Once this hyperedge is established, the graph permits the construction of the function $f : Q \to B$.

This is fine as a definition but of course equality is not always easy to prove.  Sometimes the difficulty is essential, but sometimes it is not: consider the equality between a general arithmetic expression and the natural number it equals. Within our definitions of $\St$ or $\U$ so far, it would seem that showing this requires a lengthy search with a large branching factor. But actually there is a simple algorithm to reduce any arithmetic expression to a canonical form (a natural number), making this straightforward. More generally, many deductions and calculations are straightforward, with a single preferred option for each choice. Finding such opportunities and using them is central to doing mathematics, and not yet represented in $\St$ or $\U$.

One way to represent canonicalization is by assigning a ``rank'' or ``energy'' to statements, and always following hyperedges which reduce the rank. Another way, called definitional equality,
is to mark the hyperedges with this information and prefer one side of a construction to the other. For example, the axiom $a + Sb := S(a+b)$ is always interpreted by replacing the LHS with the RHS.  This can be contrasted with $Sa + b := S(a+b)$ which is a proposition with a nontrivial proof.  See the appendix for more on these examples.

\subsection{Formalism, complexity measures and notations}
\label{ss:complex}

Now that we have sketched the hypergraphs $\St$ and $\U$, we can introduce some basic formalism for working with them and, more importantly,
define a number of universal complexity measures. These complexity measures may play an important role in determining the objective importance of various mathematical theorems, proofs, or other constructions. 

Let $\HE$ be a set of kinds of hyperedges (for example $\HE$ may contain Modus Ponens). Define the extension operator $\CT_\HE\G$ to produce the extension of $\G$ by all single hyperedges from $\HE$ with inputs in $\G$ and consistent with the constraints (e.g. of type theory).  Starting from a hypergraph $\G_0$, let $\G_{i+1}=\CT_\HE\G_i$. We define the inductive closure $\mbox{Cl}_{\CT_\HE}\G_0$ to be the union of the $\G_i$.\footnote{  It is also the least fixed point of the operator $\CT_\HE$ and is unique by the Kleene fixed point theorem given the finiteness assumptions.} It is well defined if $\G_0$ is finite and every hyperedge in $\HE$ has finitely many inputs. We define the {\bf depth }\rm  $d(s)$ of a node $s$ to be the minimal $d$ such that $s\in \G_d$. 

For our hypergraphs, we choose some minimal set of rules of construction $\HE$ and some initial elementary statements for $\St_0$, then $\St$ is the inductive closure $\mbox{Cl}_{\CT_\HE}\St_0$. Let $\Pi$ be the projection onto propositions and $\pi$ be the projection onto proofs.  We denote the proof map $\pi\St\rightarrow \Pi\St$ as $\Proof$.

Now, choose some $\hat \U_0$ to be the axioms, proofs by definition of associated propositions in $\U_0$. Then the inductive closure $\hat \U\equiv \mbox{Cl}_{\pi\circ \CT_\HE} \hat \U_0$ is the universal hypergraph of proofs, and its image $\Proof\;\hat \U$ is the universal hypergraph $\U$. The hyperedges of $\U$ representing deductions are images of those in $\hat \U$.

Define a neighborhood (or ball) of a statement $s$ as $\N(s,d) = \cup_{i\le d} \mathcal{T}^i_\HE\{s\} \subset \St$. Let $\N(s)$ (for propositions) be $\N(s,d)$ with the smallest $d$ guaranteeing that $\neg s\in\N(s)$ ($d$ will generally be small as negation is a simple construction). Then, a basic task of a mathematician is, given a proposition $s$, find the intersection $\N(s) \cap \U$, that is to determine whether $s$ or similar statements have proofs.

We can further generalize the depth function in the following way. We associate a complexity measure $c(e)$ to the edges, which can depend on the specific inputs to that edge.  We furthermore allow a term $c'(s)$ associated to the inputs, for example to account for the complexity of the abstractions it uses.
Then the complexity of a sub-hypergraph $\G$ is the sum of that of its edges and input nodes,
\begin{align}\label{eq:complexity}
    c(\G) = \sum_{e \in \G} c(e) + \sum_{s\in \G_{inputs}} c'(s).
\end{align}
This covers many cases, in particular the complexity of a definition $\G\subset\St$,
and that of a proof $\G \subset \hat\U$ (in the proof node framework) or $\G\in\U$ 
(in the framework in which $\U\subset\St$ has only deduction hyperedges). For example, suppose we set $c(e) = 1$, $c'(s) = 0$ and let $\G \subset \U$ be a proof of some proposition. Then $c(\G)$ counts the number of deductions in the proof. 
As for an expression $\bar{s}$, recall that it is associated with a sub-hypergraph which constructs that expression. Therefore the length of an expression can also be cast in terms of Eq. \ref{eq:complexity}.  Making specific choices for $c,c'$ we fix a definition $l(\bar{s})$.

We define these quantities for a node $s$ by minimizing over all sub-hypergraphs $\G^{\rightarrow s}$ which end on $s$:
\begin{align}
\label{min_complexity}
    m(s) = \min_{\G^{\rightarrow s}} c(\G^{\rightarrow s}) . 
\end{align}

Similarly we can define the length of a node $l(s)$ as the length of the smallest expression that represents $s$, to get a variant of Eq. \ref{min_complexity},\footnote{In \S \ref{ss:efficiency} we will compare these definitions with Kolmogorov complexity. } 
\begin{align}
\label{node_length}
    l(s) = \min_{\{\bar{s}\}} l(\bar{s}).
\end{align}
Shortly we will ask whether such complexity measures correlate with the importance of various structures in mathematics. 

\section{Universal properties of mathematics}
\label{s:universal}

In the preceding section, we sketched a formalism to capture the underlying structure of mathematics. The purpose of this section is to pose a number of basic questions that arise immediately from such a consideration. 

\subsection{Many Platonic worlds}
\label{ss:platonic}

Let us begin here by discussing some known facts about mathematical proof in the language of the preceding section. From Gödel's work, for consistent systems rich enough to construct self-reference, for example first order Peano Arithmetic (PA), there will inevitably be infinitely many independent statements that can neither be proved nor disproved within the system.  These simply would not appear in $\U$.  However, the very fact of a statement's independence means we are free to add it (or its negation) as an axiom and enlarge the system. So the various system hypergraphs, themselves, have a nested structure.  In the case of Peano Arithmetic, there is an accepted simplest model within Zermelo-Fraenkel (ZF) set theory, so from the perspective of ZF, we may be able to decide that a PA-independent statement is "true" by  proving it in ZF.  Using this perspective we can construct larger, extended, hypergraphs of PA by adding as axioms ZF-true sentences within PA. This process, anchored in provability can yield larger hypergraphs more closely representing "truth".      For example, there are well-known theorems of set theory (ZF) like Goodstein's theorem, which cannot be proven (according to the Kirby--Harrington theorem) within PA. The explanation of this theorem is that PA has non-standard, larger, models (within set theory)  where the Goodstein theorem is simply false, much as Euclid's parallel postulate is false in the hyperbolic plane and so cannot possibly follow from Euclid's other axioms, which hold in both Euclidean and hyperbolic contexts. In PA, the infinite set of true (provable in ZF) statements that are not provable from the axioms is not \textbf{even} computably enumerable \cite{Enayat2014StandardModels}. For this reason, the totality are not considered to be axioms -- if they were, then it would not be possible to computably check proofs because a Turing machine could not determine whether any given statement is an axiom.  

So this highlights a choice, if we are so ambitious to try to model "truth" we will not even know all our starting points. We are happy, for now, to restrict attention to "provable" from some standard set of axioms and rules: PA, ZF, ZFC, or even some tiny decidable system like propositional calculus (PC).
Of course, given that we can consider formal systems with various axiom systems, different symbolic languages, and different deductive rules, there are an infinite number of mathematical worlds (planets to explore), each with its own hypergraph and additional structure as described above. We will refer to each of these hypergraphs and their associated structures as a Platonic world. 

\subsubsection{Coarse equivalence}

Two proof systems can share all of their provable statements, even though the geometry of their hypergraph of proofs may look quite different between the two.
An ``absolute'' statement is one which is provable across both, or many logical systems.  For example, ``$2+2=4$'' is true in almost all formal systems which contain arithmetic.
Any statement in (first order) PA which is provable in ZFC is also provable in ZF, accordingly PA is said to be absolute from ZFC to ZF.  In this case there is a polynomial time algorithm to eliminate choice, so if one imagines, at the hypergraph level, the inclusion PA $\subset$ ZF and PA $\subset$ ZFC we might say the embeddings are "coarsely\footnote{Our use of the word "coarse" is a nod to the subject of coarse geometry \cite{wikipedia_coarse_structure} a concept well explored by M. Gromov  and his school. In their context, "coarse equivalence" is usually base on bounded multiplicative distortion of length. Proof theory is a wilder context and polynomial distortions (at least) should likely be allowed.} similar". Curiously, there are examples of absoluteness in higher arithmetics where there the elimination procedure is so explosive in growth as not to be recursive \cite{Pudlak1998LengthsOfProofs,Buss1994GodelLengthsI,Fischer2014TruthAndSpeedup}. This is a warning that even coarse geometrical features of  "proof graphs" may depend on the choice of foundations. Thus our meta-mathematical context is quite unlike many familiar mathematical contexts, e.g. geometric group theory, where coarse notions such as the "growth of a group" are independent of the choice of generators. In math, what can be done quickly does depend on the choice of formalism. A future project might be to extract some "foundation-independent" picture of mathematics.

At this point we encounter a theme espoused by Hugh Woodin \cite{Woodin2001CHPart1} (and before him Gödel \cite{Godel1947CantorContinuum}), where both speak of finding the "right" and "fruitful" axioms in cases where formal independence has been established, but, in their view, does not constitute the final word. The apparent diversity of mathematical worlds based on axiom choices and foundational formalism (set theory, type theory, category theory, topos theory) may at some stage evaporate as a consensus on the "best" or "correct" foundations emerges.  We do not yet have sufficient perspective to know if the "Worlds" in the title of this essay will eventually  become a singular noun.

In light of the above discussion, let $\UU$ denote the collection of all universal proof hypergraphs of formal mathematical systems and let $\U \in \UU$ denote a particular system hypergraph. Within any given formal system, the sum total of human knowledge of mathematics fills out a subgraph $\Hum \subset \U$, with $\HH$ denoting the collection of all of humanity's mathematical knowledge (expressible) in different formal systems. Often the "same" fact will be expressible in many different systems.  Comparing how easily it is obtained across these systems is data useful for comparing the systems.

\subsection{Questions about the nature of Mathematics}

We can now ask structural questions about the nature of Mathematics.
    
\begin{enumerate}

    \item {\bf How much depends on the formal system?} Are there just a few universality classes of mathematical systems in terms of the "coarse geometry" of their proof hypergraphs? One might expect all sufficiently strong systems to lie in one class. This question touches on absoluteness and on the efficiency of rewriting results in one system to another, which can vary from polynomial to non-recursive.  

  \item  {\bf What about the "geometry" of the hypergraphs in $\UU$?} Here we have committed a slight abuse of meaning. Geometry, with all its embellishments, is based on a simple binary relation, distance. Our use of the term in quotes anticipates some bold generalization of geometry - replacing the binary distance with the n-ary structure of hyperedges - permitting a rigorous discussion of the "geometry of math".  Does Gromov's combinatorial notion of curvature have a hypergraph analog? What kinds of scaling limits can be obtained -- do manifolds, Euclidean or Lorentzian ever emerge?
  
  \item {\bf Is there a graph theoretic, geometric, or perhaps information theoretic way to pick out subgraphs $\HH$ of human mathematics from the formal system's hypergraphs $\UU$ containing it?} In \cite{AksenovBodniaFreedmanMulligan_ToyModels} evidence is presented that the human sector, called HM therein, has merely polynomial growth, and is for that reason called a "ribbon". This question has two aspects: What is the intrinsic structure of  $\HH$, and what is the structure of the pair ($\HH$, $\UU$).
\end{enumerate}

\subsection{Modeling worlds $\U$ and world pairs $(\U, \mathcal{H})$}

\iffalse
Gödel, Tarski, and their successors have  produced some understanding of both provability and of truth within a model.  One can think of Gödel's results as illuminating the posible ``lower'' \it boundaries \rm of the (truth) hypergraphs in $\UU$. For example, demonstrating that there are potentially an infinite number of root nodes (that do not even form a computably enumerable set). On the other hand, the structure in the \it bulk \rm of the hypergraphs seems to have escaped the attention of logicians and philosophers of mathematics, although one can argue that complexity theory is an attempt to address this bulk geometry \cite{aaronson101philosophers}. \maissam{Is the preceding statement too strong?}
\fi

Gödel, Tarski, and their successors have produced deep insights into both provability and truth within a model. One can think of their foundational results as illuminating the possible ``lower'' boundaries of the truth hypergraphs in $\UU$—-demonstrating, for instance, that there must exist an infinite, non-computably enumerable set of root nodes. Traditional philosophy of mathematics has historically fixated on these boundaries. The rich structure and geometry within the bulk of the hypergraph has largely been left to structural proof theorists, category theorists, and computer scientists, where fields like computational complexity actively attempt to quantify this bulk geometry \cite{aaronson101philosophers}.

We expect that AIs will reveal new statistical information about the bulk, which could suggest statistical models of provability and help to fit their parameters. AIs open the possibility of building toy models for how mathematics grows out from its roots. The experience from condensed matter physics is that even the simplest toy models, particularly if exactly solvable, can yield deep insights into messy real-world problems \cite{baxter2013exactly}. A metaphor for the kind of parameter fixing we anticipate is SLE (Schramm - Loewner evolution) \cite{Wikipedia_SLE_2025_oldid1318347270}. There a continuous family of conformally invariant 2D models (including: loop-erased random walks, self-avoiding random walks, critical Ising interfaces, Gaussian free fields, etc...) each result from some specific value of a Brownian drive parameter $\kappa$ in one real dimension. Knowing the fractal dimension of the random structures allows one to solve for $\kappa$.

Graph theory has matured significantly as a subject over the last few decades.
Graphs can be analyzed mathematically and quantitatively, in terms of clusters, bottlenecks, motifs, expansion properties, spectral properties, small-world and scale-free natures, and so on. It seems natural now to search for the structural properties of  the hypergraphs in $\UU$.\footnote{
Works studying the network structure of ITP libraries include \cite{huch2022formal,viteri_epistemic_2022,bauer2023mlfmf}.
} Is it an impossibly dense, infinite homogenous shrub of \textbf{Truth} (or more exactly \textbf{proof}), or are there observable absolute structures lying within the hypergraphs in $\UU$?

Although it will be impossible to fully grasp these Platonic worlds, perhaps we may learn more about their intrinsic structure by designing AIs to traverse them. We are particularly interested in an intermediate ``mesoscale.'' Note that what counts as mesoscale to a human and an AI may differ by a factor of a million, but this is a small number set against the combinatorial explosion of formal mathematics; we and our AIs both need to explore parsimoniously.

The fine structure of  mathematics, human or otherwise, is a bit boring, and the largest scales can at best be lightly sketched, but at the right altitude, at some mesoscale,  we hope and expect that something akin to geometry emerges and that we can learn the features of a new landscape.  Human-led mathematics has painstakingly searched for structures in both  $\HH$ and $\UU$. AIs will soon be tasked with discovering them for us. As they do particular attention will be paid to the relative ``geometry'' of the pair $(\UU,\HH)$. AIs must learn to notice landmarks which will keep them on, or near, the human trail of exploration, and guide autonomous exploration in human-like and human-comprehensible directions.

Our minds impose an ``interestingness'' prior over $\UU$ -- certainly only a tiny fraction of correct deductions, even if fully parsed, would seem interesting to us. There may be an objective, algorithmically computable, term in such priors as well as more environmental ones.\footnote{We note that research in automated mathematical discovery dates back decades \cite{wang1960,Lenat1977IJCAI}, and hand-crafted notions of interestingness have been proposed and used in various contexts \cite{ColtonBundyWalsh2000Interestingness}; here our primary discussion is in the context of the proof hypergraphs $\U$ and $\Hum$.} The ratios discussed below are one objective candidate, another is the degree to which a formal statement compresses. Compressibility is  amenable to rigorous definition and can be autonomously explored. 

Our priors are also influenced by our human experience as biological and social creatures evolved in our specific physical universe.  These change in time through social dynamics and feedback from science and engineering, and could take longer to transmit to bots.  Let us save them for later discussion and here explore more objective measures of  importance.
While $\mathscr{U}$ appears ferociously complicated, there is a long history of large universal objects, like classifying spaces, turning out to have important simple features, like Bott periodicity.  We are hopeful that someone (or some AI) will someday say something smart about $\mathscr{U}$.

\subsection{Universal importance measures: Theorems and willow trees}
\label{sec:universal_importance}

In mathematics we have a ranking of the importance of proved statements, and ``theorem'' is generally used for statements toward the top of the pecking order.\footnote{
Some famous linguistic exceptions are Dehn's Lemma and Schur's Lemma.} 
What are the origins of this ranking? There are at least three places to look:\footnote{For simplicity we use graph rather than hypergraph terminology.}
\begin{enumerate}
    \item There are objective graph-theoretic reasons for a particular node or "path" (proof) in $\U$ to stand out, independent of our human environment. 
    \item A path has special graph-theoretic structure in $\Hum$, but not necessarily in $\U$. 
    \item Human-centered reasons, sociological or otherwise. This could be distinct from (2) if our prior were largely Human-centered.
\end{enumerate}

\subsubsection{Efficiency and length vs. proof complexity}
\label{ss:efficiency}

Here we will introduce a measure of \it efficiency \rm $E(\bar{P})$ of a proposition $\bar{P}$ as one possible measure of its importance or interestingness.  If a complicated proof can be avoided by quoting a simply stated theorem, that is clearly an efficient thing to do.

To formalize this, we define the complexity of a proof $\bar{p}$ of $\bar{P}$, to be the total length of all statements along the proof:
\begin{align}
    c(\bar{p}) = \sum_{\bar{s} \in \bar{p}} l(\bar{s}).
\end{align}
We also have the minimum complexity, which minimizes over all proofs $\bar p$ of $\bar{P}$,\footnote{This is uncomputable if we minimize over $\U$ but computable over $\U_t$, which is one reason that interestingness is contingent. Similarly we minimize Eq. \ref{node_length} over $\bar P\in\St_t$ below.}
\begin{align}
    m(\bar{P}) = \min_p c(p) .
\end{align}
Thus we define the {\it efficiency} of $\bar{P}$ to be
\begin{align} \label{eq:efficiency}
    E({\bar{P}}) := \frac{m({\bar{P}})}{l({\bar{P}})}.
\end{align}
This was defined in terms of $\bar{p} : \bar{P}$ written in the symbolic language.  When we go to the hypergraph we can use a similar definition in terms of $p : P$, but now there is an additional possibility that many statements $\bar{P_i}$ and corresponding sub-hypergraphs construct the same proposition $P$.  Thus we should also minimize over this choice and use $l(P)$ as defined in Eq. \ref{node_length}. This reflects the importance of ``definition'' and ``lemma'' in mathematics, which enable us to shorten statements by using definitions.

Going to extremes, the ultimate compression of any string is its Kolmogorov complexity, so set $l'(\bar{P})$, $m'(\bar{P})$ to be the Kolmogorov complexity $C$ of $\bar{P}$ and the least complex proof of $\bar{P}$ respectively, and define $E'(\bar{P}):= m'(\bar{P})/l'(\bar{P})$.\footnote{Clearly graphs in any $\U\in\UU$ constructing $p$ and $P$ are possible minima for Kolmogorov complexity, so Kolmogorov complexity only differs from our definition if there is some unknown formal system $\U\in\UU$ which greatly shortens statements and proofs.} All versions of efficiency have their pro's and con's, and are enumerated mainly to stimulate thought on how to describe "efficiency", "importance", and "interest" for our AI friends; see \cite{AksenovBodniaFreedmanMulligan_ToyModels} for a mathematical toy model for definitions within human mathematics, and other starting points for the concept of mathematical interest.

These measures of efficiency merely scratch the surface: there is also `logical depth' proposed by Bennett \cite{bennett1988logical}, which emphasizes\textbf{ run-time} of the shortest program (or at least relatively short programs) that generates a string, and is a prescient philosophical introduction to what makes patterns remarkable in nature and mathematics. Bennett thinks of depth as a measure of the "mathematical work" needed to build a structure. This concept seems attractive for measuring $m$, less so for $l$, so the best notion for $E$ might necessitate mixing and matching. 

If we treat $l$ as a height function on the hypergraphs, and taking efficiency to be a measure of importance, we can imagine (with some artistic liberty) what mathematics may look like. The hypergraph is tree-like, so take the liberty of imagining the height function literally.  We are now looking at a beautiful willow tree, the willow tree of mathematics. The height function of each vertex defines its height above the ground.  Important theorems correspond to leaves hanging near the ground, coming down from branches that ascend into the sky before coming back down. We can further imagine reaching these leaves by crawling up and along branches, perhaps all the while fearing to get too high (complex). We like to think of our mathematical selves running through the grass and leaping up to touch the lower leaves: mathematicians have always been children at heart. Maybe mathematical frameworks can be likened to ladders which enable us to climb high up at particular places.

Other factors are also likely important in defining the complexity of a proof and using it to define importance measures. For example, whether the proof goes through a bottleneck in the graph. Or to what extent  there are neighboring proofs, allowing the proof to be perturbed.\footnote{This touches on Hilbert's 24th problem \cite{Thiele01012003}, which asked for a theory of proofs, to prove the simplicity of proofs and studying deformations between different proofs.} It would be an interesting endeavor to quantify these complexity measures and study them in practice. 

Note that our definitions depend on the choice of hypergraphs $\U_t$ and $\St_t$ to minimize over.  We can just as well apply them to human mathematics $\Hum_t$.
We may also apply them to the full hypergraphs $\U$, $\St$ in any given formal system; in this case the complexities and efficiencies discussed above are uncomputable, but they still may be of value, analogously to the notion of Kolmogorov complexity. 

\subsubsection{Hubs and bottlenecks}
\label{ss:hubs}

The importance of a statement could well be correlated with the node's connectivity properties. For example, an important statement corresponds to a node in the hypergraph that acts as a hub, with an outsize number of outward-directed edges that eventually flow to other important statements. Perhaps a person or an AI might see something about the structure of the graph, e.g. some cluster properties, enjoyed by that node. This is reminiscent of a recent discussion in \cite{BengioMalkin2024AItheory}, which suggests to view the importance of theorems from the perspective of compression: an important theorem shortens the proof of other theorems. 

Alternatively, the node might be part of a crucial bottleneck that connects different clusters of the hypergraph, corresponding to connecting different fields, as in the proof of Fermat's last theorem. 

Analyzing connectivity properties for $\U$ is difficult since each node in principle has an infinite number of connections. To make progress, we would need to define a measure on the nodes of $\U$ which decays to zero fast enough with $l(s)$, $d(s)$, or other measure of complexity of $s$, beyond a finite cutoff. Then the connectivity of each hyperedge would have to be weighted against this measure to give a finite value. Since the growth is at least doubly exponential, it is not clear that such measures can be appropriately defined and are distinct from a finite cutoff. 

In this view, there is something about the structure of the hypergraphs, either $\Hum$ or $\U$  or the pair $(\U,\Hum)$ and the length (or entropy) of statements that singles out certain statements as theorems. It might be that the importance of the statement only comes from the node and its graph-theoretic properties within $\Hum$, but that as a node in $\U$ it is unremarkable. Or, it might be that the node has remarkable graph theoretic structure in an absolute sense within $\U$. In the former case, the importance of the theorem is due to its relation to currently known human mathematics, whereas in the latter case, there is an absolute sense of its importance.  (In this discussion we have loosely appropriated graph-terms to hypergraphs, we have only begun to explore the apt generalizations.)

There will be analogies, relationships, motifs, between similar structures from different Platonic worlds, and so focusing on the hypergraph of a single Platonic world is not sufficient to be able to fully mark the importance of a given statement. A statement may be important because many Platonic worlds have a similar statement.

\section{Models of mathematical thought}
\label{s:models}

In \S \ref{s:universal} we took the point of view that mathematics has a universal structure which is being gradually discovered by humans, soon with the help of AI. While this is a time-honored position, there is an alternate point of view: mathematics is a collection of tools for thinking which are invented. This point of view suggests different questions: What are these tools and how are they used? What determines which tools are more useful, and more effective at the tasks of mathematics?

It also suggests a different way to get at the question of what determines $\Hum$. We can propose a {\bf model} of the process of creating mathematics which idealizes the human process, and study it to gain insight. This process includes the formulation of new definitions and conjectures, and the steps in proving a conjecture or finding a counterexample.  It also includes making a decision for each new result: Is it important enough to remember and publish?   Or on the other hand, is it a result which could be easily re-proven from other more significant or more interesting results, in which case the cost of remembering it might outweigh the benefit.  This is the point addressed by defining an ``interestingness'' measure.
These are central questions for mathematical discovery -- while we restrict to this topic here,
one could also model other parts of the process of doing mathematics which could bear on the structure of $\Hum$: formulating and solving problems, teaching and communication, {\it etc.}.

Using the hypergraphic logical framework we described, our model will be a process which generates a sequence of sub-hypergraphs $\C_t \subset \St$ for $t=0,1,2,\ldots$, in which each time step adds or removes a single hyperedge or a single definition. We will discuss such process models shortly, and here ask:
\begin{enumerate}\setcounter{enumi}{3}

    \item {\bf How can we model the agents (human and AI) which explore and construct parts of $\UU$?}  A central goal of computer science is to develop and understand algorithms for proof and for the other tasks of mathematics (discovery, search, communication, {\it etc.})
    Are algorithms a good way to think about the human process, or that of the society of human mathematicians?  Is there universality with respect to algorithm? As we discussed earlier, complexity of proof calculi is not universal. One can ask if all ``sufficiently powerful'' algorithms for finding proofs are comparable.\footnote{Looking at human mathematicians, the contributions of Gromov and Deligne, two comparably powerful proof generating algorithms, have little overlap in methods or results. However the corpus of  von Neumann or  Milnor  gives a more universal impression.} 
\end{enumerate}

What will come out of such a model?  While the details of human mathematics at any given time $\Hum_t$ are surely too complicated to reproduce, its general properties might emerge from simulations and analyses of simplified versions of the model. Even these dynamics will be complicated, but we can cut through this by formulating the model in terms of an objective function which the dynamics tries to optimize.  An appropriate objective function might be motivated by considerations natural from the agent point of view (increasing its efficiency to gain rewards), and the optimal hypergraphs $\Hum_t$ might turn out to be selected (in a non-obvious way) in terms of natural graph-theoretic properties such as the complexity measures of \S \ref{s:hyper}. In this way, answering question 4 might  answer question 3.

\subsection{General nature of the models}
\label{ss:models}

We consider models -- call them ``agents'' -- formulated in the language of reinforcement learning (RL) \cite{sutton1998reinforcement}. An agent is a process which evolves a sequence of states $S_t$ through successive times $t=0,1,\ldots$. At time $t$ the agent chooses an action $a_t$ which then leads to a transition $(S_t,a_t)\rightarrow S_{t+1}$.\footnote{ In general this transition is stochastic, with a probability depending on $S_t$ and $a_t$. The standard terminology is ``Markov decision process'' (MDP).} A transition also comes with a reward $R_{t+1}$, a real-valued function of $(S_t,a_t,S_{t+1})$. This combination of data (state space, set of actions, transition function, rewards) is called the environment. The agent's objective is to maximize the expected discounted future reward.  

The agent starts out only knowing the set of actions and perhaps general properties of the state space. It does not know the actual state space, or the transitions or rewards.  Rather, it must learn these by exploring the environment and gaining rewards.\footnote{In practice the agent is usually allowed to observe other properties of the state, for example the board position in chess.  But unlike chess, in most tasks (and ours) the state is too large to completely observe. } While how it does this is open-ended, the central thing it learns is a policy function by which it chooses the next action $a_t$. A standard definition here is to choose the policy function to come from a parameterized class of functions $\pi_\theta(S_t,a_t) \in [0,1]$, whose
value is the probability that the agent will take action $a_t$.  The agent then learns 
the parameters $\theta_t$ by some optimization procedure based on its past rewards.

In mathematical discovery, the state $S_t$ includes a set of ``known'' nodes (propositions, proofs, abstractions) $\C_t \subset \St$.\footnote{This is granting the ``proof node'' formalism of \S \ref{ss:proofobj} in which proven propositions are nodes with associated proof nodes.
In another formalism the state might distinguish proven nodes $\T_t \subset \U$ and the other known nodes $\C_t \subset \St$.} The actions modify the state by adding or removing nodes and hyperedges from $\C_t$. The agent can also have internal state (heuristics, intuitions, AI weights) which we will not make explicit.

The AI models we will discuss all carry out the following iterative loop:
\medskip
\begin{tcolorbox}[title=Generic discovery agent (schematic; cf.\ \S\ref{ss:models})]
\textbf{State:} $S_t=(\C_t,\theta_t)$ where $\C_t\subset\St$ is the current corpus (nodes + hyperedges) and $\theta_t$ collects learned parameters (LM weights, heuristics, value functions, etc.).\par\medskip

\textbf{(1) Goal generation:} Pick a theorem to prove or a problem to solve. This could
be given from the outside, or it could be generated based on the current state in some non-deductive and probabilistic way.\par\medskip

\textbf{(2) Attempt:} Try to prove the theorem or solve the problem, placing some limit on the time spent trying. If this succeeds, remember the successful proof or arguments. If this does not succeed, extract partial results and remember the arguments leading up to them.\par\medskip

\textbf{(3) Learning \& abstraction:} Learn from these results in various ways. One way is AI model training (usually by updating weights): a prover can learn to prove better, a conjecture generator can learn to make better conjectures. Another is to create new abstractions -- if a particular proof or proof component has been used many times, it is a good candidate for abstraction. Similarly for functions, data types and so on.\par\medskip

\textbf{(4) Curation/compression:} Decide which of the new nodes (conjecture, proof, abstraction) to add to $\C_t$ to obtain $\C_{t+1}$. If nodes in $\C$ come with a cost, the agent might also remove nodes. The combination of addition of abstractions followed by removal of
redundant nodes is compression of the hypergraph.
\end{tcolorbox}
\medskip

The agent repeats these steps {\it ad infinitum} to define a process of mathematical exploration and creation, in which its discoveries are recorded in the sub-hypergraphs $\C_t$. The acid test -- not yet achieved as far as we know -- is for it to continue to make new and interesting discoveries for as long as it is run.

Some issues which come up in doing this:
\begin{itemize}
    \item How to generate conjectures/problems of the right level of difficulty: solvable but not too easy.
    Heuristics such as induction from examples and unsound rules of deduction are helpful.
    \item Brute force search for proofs and solutions is very limited.
    This type of search is like game playing and one can apply successful methods there.  Still, one inevitably
    hits the barrier of exponential complexity.  The only solution to this is to make the proofs shorter.
    \item Abstractions can drastically shorten proofs, but they are not easy to find and have costs,
    of remembering them and of increasing the branching factor.
    \item Unlike most tasks in AI, it is not known what objective function to optimize.
    The process of finding individual proofs can be analogized to game playing and success can be used as a
    signal for reinforcement learning.  But attributing rewards to creating abstractions and to exploration
    is still an open research problem.
    
\end{itemize}

\subsection{Making conjectures}
\label{sec:conjecture}

The simplest way to produce sensible conjectures is by using {\bf unsound} rules of deduction.  These could be purely syntactic rules: for example, if $A \Rightarrow B$ then conjecture $B \Rightarrow A$.  They are similar to the sound rules of deduction we have been using so far, the difference is that they do not preserve truth. But, when used in the right contexts, they are powerful heuristics whose outputs have a much higher probability of being true than random
propositions in $\St$.\footnote{A related idea, popular in AI, is probabilistic reasoning.  This allows working with unproven statements while mitigating the problem that contradictions are fatal to classical logic.}

A more systematic method is inductive generalization.  In general terms, this is based on a population and samples from the population. Given enough diverse samples, if a property $P$ holds for all of them, one is motivated to conjecture that $P$ is true of the entire population.  For us, the population will be objects in a class as defined in \S \ref{ss:equality}: abstract groups, manifolds, and so on.  The use of this heuristic motivates the collection and curation of examples (curation means keeping a variety of interesting ones and discarding others).  
It also requires coming up with candidate properties $P$. These might arise in the course of proving lemmas about the class, or they might be generated systematically.  

A closely related method is function fitting -- we are given many input-output pairs and asked to
find a simple function fitting them.   While not always thought of as pure math, this is a crucial type of conjecturing. Fitting functions with symbolic expressions and/or programs is a classic machine learning topic, and both forms of inductive generalization appear in the AI math discovery systems we discuss in \S \ref{ss:agentic}.

There is a mystery here: $\U$ does not contain conjectures, and one can imagine an ultra-powerful AI mathematician (an ``oracle'') which can answer very hard questions with a proof or disproof just based on its knowledge of $\U$. But in human mathematics and all of the AI systems we will discuss, conjectures are very important. This raises the question
\begin{enumerate}\setcounter{enumi}{4}
\item {\bf Can one do mathematics without conjectures?  If not, why not?}
\end{enumerate}

This may be related to the final point, the lack of a clear objective function. Objective functions measuring performance are a key element both in human and in machine learning. For RL one uses the expected discounted future reward. For playing a game or proving a specified theorem, the reward is specified in advance and from the outside: these are ``extrinsic'' rewards.  But much progress in mathematics involves finding both the statement and the
proof of a theorem, or both a task and its solution. Taking a very general point of view, one says that  such a system is working with ``intrinsic'' rewards \cite{chentanez2004intrinsically}.  The simplest example is to reward the agent each time it reaches a new state.  This favors exploration and works well for (say) learning to solve a maze, but not for mathematics -- only certain new states are ``interesting.'' There are more sophisticated measures of intrinsic reward which we will touch on below. 

One could try to argue that ultimately, all of the rewards are derived from extrinsic rewards.
This seems reasonable for problems coming out of scientific or practical applications, and one could model it by putting the agent in an environment which requires performing idealizations of these tasks. But another idealization, more intrinsic to mathematics, is to choose a simple and natural class of pure math problems, and base the reward on the ability to solve them.  For example, one can stake out a sizable portion of number theory as the task: given a Diophantine equation, show how to find its solutions  or argue that doing this is intrinsically hard.  
A verifiable proof of one of these alternatives would count as ``resolving'' the  problem.
This task can be made precise by postulating a distribution over the equations. Sampling from this distribution provides an unlimited ``benchmark'' of problems whose resolutions can serve as rewards.

This idea can be generalized by framing ``solving equations'' as a ``universally interesting'' mathematical task, leading us to ask what are other universally interesting tasks. One is to, given a signature-axiom class, classify the instances up to isomorphism. This task is particularly emphasized by McAllester \cite{mcallester_mathzero_2020} as central to mathematical thought. Another is whether a set with a simple definition ({\it e.g.} primes) is empty, finite or infinite; and if its elements have a natural size then approximate the corresponding distribution over sizes.

\subsection{Finding proofs}
\label{s:findproofs}

Given $\C_t$ and a conjecture $s$ (and possibly additional hypotheses or variables), the agent searches for a larger sound hypergraph in which $s$ is proven, $s\in\C'\cap\U$ with $\C'\supset \C_t$ (proving $\neg s$ would also count as success). A simplified picture of this is in terms of a tree with a branching factor $b$, the number of choices at each level of the tree, and a depth $D$ at which the goal appears.  Very roughly one can estimate the time required to find the proof as $T \sim b^D$. The naive search tree (including all allowed hyperedges) is intractable, so real provers use heuristics to reduce the search space, by choosing a small number of nodes to consider as inputs to the new hyperedges (this is called premise selection), and by pruning redundant or unpromising branches. This reduces $b$ at the potential cost of removing correct arguments and thus increasing $D$.

Even with these heuristics, the combinatorial explosion of $\U$ means that brute force search is not an effective way to find long proofs. This is a central point of theoretical computer science, formulated in the $\P \ne \NP$ conjecture and other separations of complexity classes.  Nevertheless humans can find long proofs. Is this a selection effect: we find the proofs which are easy to find?  Or are there clever ways to prove the statements which arise in practice?
Surely both are true, but let us focus here on the second point.

\subsubsection{Forward and backward chaining}

Two broad strategies for proof are forward chaining and backward chaining. In forward chaining, one works with proven statements and uses rules of deduction to prove new statements. This can be modeled by extending our knowledge $\C_t \cap \U$ to successively larger subgraphs of $\U$.  While this is guaranteed to produce only true statements, most statements deduced this way are not useful for proving $s$, and it is difficult to know in advance which ones are useful.

In backward chaining, one works backwards from the goal $s$.  A backwards reasoning step postulates a finite set of new statements $\{t_1,\ldots,t_k\}$ such that it is easy to find a proof of $t_1 \wedge \ldots t_k \rightarrow s$.  This has the advantage of being more likely
to be useful for proving $s$ if the $t_i$'s can be proven, and the disadvantage that one might choose $t_i$ which are incorrect or too difficult to prove.  

Both strategies have advantages and ITP systems generally use both.  A strategy which uses both
is  called ``meet in the middle.'' Suppose the shortest proof is $D$ steps long, then a brute force search will take time $b^D$. But if we pursue the search in both directions, we remember all the intermediate results and we can tell when both searches obtain the same result,
then a proof can be found in time $2 b^{D/2}$.  This idea can be extended to use more intermediate steps: one can envision a ``hierarchical planning'' proof strategy which works by finding (say) every tenth intermediate step of the proof and then filling in the details.  This would reduce $b^D$ to $C b^{D/10}$. The canonicalization theme of \S \ref{ss:equality} leads to further valuable methods for reducing the branching factor.

Taking into account these various strategies, which change the branching factor and exponents but do not change the exponential complexity of search, one can suggest the following simplified picture: the time $T$ to prove a statement $s$ is $T \sim b_{\text{eff}}^{\alpha m(s)}$, where $b_{\text{eff}}$ is an effective branching factor which in a good system is not too much larger than $1$, $m(s)$ is the complexity (length) of the shortest proof and $\alpha$ is an empirical factor also of order one.  This simple model and realistic constraints on $T$ suggest that there is a maximal complexity $m_{\text{max}} \sim \log T_{\text{max}} / \alpha\log b_{\text{eff}}$ below which proofs can generally be found and beyond which this becomes very unlikely. 
The main avenue for going beyond this complexity barrier is to reduce $m(s)$ by the use of abstractions.

\subsubsection{Linearization}

Another way that proofs may simplify is through ``linearized" deductions. Consider a fixed hypergraph $\G\subset\St$, which effectively serves as a fixed library, and a general statement corresponding to a node $s \in \St$. We define a linearized deduction as one which can be made by connecting a series of hyperedges $s_1 = e_1(g,g',\ldots,s)$, $s_2 = e_2(g'',g''',\ldots,s_1)$ and so on in which each new statement appears once as an input and the other inputs all come from $\G$. All of the inputs from $\G$ can be chosen independently, both those at different steps and the $p-1$ inputs of a $(p,1)$-hyperedge with $p>2$.  This leads to a tree structure where the branching is the choice of hyperedge and elements from $\G$. In this case, since $\G$ is fixed, the growth is only singly exponential, leading to a much more constrained search problem. It would be interesting to see if many real proofs can be modeled by networks in which most of the deductions take this form, with occasional interactions where several general statements are combined. Mathematicians are rightly impressed when disparate fields meet, suggesting "non-linearity" is not the norm.

\subsection{Finding and evaluating abstractions}
\label{s:findabs}

Abstractions are fundamental to mathematics -- a good abstraction isolates a recurring concept, which may require a large hypergraph to describe, and converts it into a small hypergraph. 
This means that AIs must not just traverse the hypergraph and look for structure, they must constantly be on the lookout for good abstractions. How will they know a good abstraction?

One possibility is that good choices lead to good graph-theoretic properties. For example, a good abstraction may be one that dramatically reduces the complexity of a large number of high interest nodes. Another possibility is that good definitions and concepts are best determined by considering many different hypergraphs simultaneously. Or maybe they require some completely external input mirroring our understanding of the physical world, and that we gravitate toward them because they systematize our experience.  Still another idea, pursued in \cite{AksenovBodniaFreedmanMulligan_ToyModels}, is to base the choices of abstractions on informational compression and Kolmogorov complexity.

Different reasoning agents will find different sets of abstractions useful. This is visible in ITP systems, whose libraries contain many definitions and proofs which humans find trivial.  Conversely a superhuman AI mathematician may feel that a two page explanation of (say) Wiles' proof of Fermat's last theorem is perfectly adequate.

As discussed in \S \ref{s:hyper}, abstractions summarize sub-hypergraphs of $\St$ and $\U$ into single nodes, and can be used to replace long proofs and definitions with sub-hypergraphs with $\CO(1)$ hyperedges.  Thus they can drastically shorten proofs; however the number of possible abstractions is also combinatorially large and only a negligibly small fraction of the possibilities can be used in practice.  How does one find a small yet useful set of abstractions?%
\footnote{One could also consider ``soft abstractions,'' meaning common patterns which have not yet been formalized (so are not accessible to the abstraction mechanisms we defined) yet which can be learned by an agent and which speed up its search. To give an example from functional analysis and non-linear PDEs, it is difficult for Cliff Taubes to write a paper less than 100 pages long, whereas in algebra 10 pages will often do. It seems that in analysis the ideal of a self-contained structure of reusable lemmas is harder to achieve than in algebra. }
Can we estimate what effect these will have on $\St$ and on $\U$?

This general question is studied in the theory of proof complexity, for example in the difference between Frege and Extended Frege systems. A Frege system formalizes textbook propositional calculus. Extended Frege adds an extention rule that allows the use of abbreviations (definitions); it remains an open problem whether Frege can polynomially simulate Extended Frege, or equivalently whether Extended Frege gives super-polynomial proof-size speedups over Frege.  We refer to the literature for further discussion \cite{wigderson2019mathematics}.

Some existing methods, such as the corpus-guided top-down synthesis approach of \cite{bowers_top-down_2023}, systematically search a corpus of knowledge $\mathcal{C}$ for sub-(hyper)graphs that can be abstracted. A plausible speculation based on such methods is that finding abstractions of size $c(A)$ in a corpus $\C$ of size $N$ takes time $T \sim N b_{\text{eff}}^{c(A)}$. This is because the abstraction is built incrementally, so at each step there is a branching factor that sets the number of possibilities. $b_{\text{eff}}$ is an effective branching factor for this process, and the factor of $N$ arises because the abstraction needs to be compared against each part of the corpus. $b_{\text{eff}}$ is definitional, not proof theoretic, as one does not have to reproduce the details of the proof in the abstraction. The largest abstraction one can reach is then set by the maximum computation time allowed, $c_\text{max}(A) \sim \log T_{\text{max}} - \log N$. Reducing this $\log N$ contribution may be one reason to divide up the corpus into topics and do premise selection. 

\subsubsection{Nested abstractions}

If we consider nested abstractions, then at each layer the abstraction could have a small size, and thus efficient to find, but when unrolled it could be exponentially long. In \cite{AksenovBodniaFreedmanMulligan_ToyModels}, the thesis is developed that all of human mathematics $\Hum$ has followed the pattern pioneered by "place notation," that is the recursive use of nested definitions to achieve exponential (and in some case greater!) compression of mathematical statements and proofs.  Place notation for numbers reduces the length $n$ definitional complexity of $S^n0$ in naive PA to $\CO(\log n)$.  In \cite{AksenovBodniaFreedmanMulligan_ToyModels} evidence is presented to support a proposed "compressibility" dichotomy between $\U$ and $\Hum$: Complexity theory limits the amount of compression generally available in $\U$ to  polynomial, whereas $\Hum$ admits, quite broadly, exponential compression. One empirical finding is that that the longest proof in MathLib (as of Oct. 17th 2025) expands from about 600 lines  to $10^{104}$ when written in tree-form using only basic Lean4 terms, an expansion by a factor of over a googol. Theorems about unstructured instances of \NP-hard problems, such as "This long CDNF Boolean formula has (or does not have) a solution"  are regarded by us as lying in $\U \backslash \Hum$ . Proofs of such statements are not compressible and are not typical of what humans can understand or care about. In \cite{AksenovBodniaFreedmanMulligan_ToyModels} the theme is developed that polynomially growing algebraic structures (in particular finitely presented monoids) admit exponential definitional compression whereas exponentially growing structures typically admit only linear compression. It is proposed that as our agents explore mathematics they should actively asses the compressibility of the terrain they encounter and seek out compressible developments, as these are most likely to be fruitful and explicable.  Even our agents, perhaps a billion times faster than us, will need to hew to compressible paths, they will be no more capable than we of of dealing with googol-long proofs.

In practice we have many examples of nested definitions. One favorite example is the sequence: magma, semigroup, monoid, group, abelian group, integers. Another entirely different kind of sequence would be $\{\mathbb{N}, \mathbb{Z}, \mathbb{Q}, \mathbb{R}, \mathbb{R}^n$, n-manifold, vector bundle, differential operator, symbol, ellipticity, elliptic bootstrap$\}$. It would be interesting to examine different sequences of nested definitions in human mathematics and try to classify their general structure and develop a meta-theory of nested definitions. 

\subsubsection{Conjecture vs. Abstraction}
\label{sec:conj_vs_abs}

There is an interesting relationship between abstraction and conjecture. Abstraction identifies recurring structure, for example frequent sub-hypergraphs, and in some cases expresses it as a predicate $P$ on objects $e$. One may arrive at $P$ by first proving $P(e_1), P(e_2), \cdots$ for particular examples $e_i$, and then formulating the abstraction $P(e)$. Conjecture formation then consists in proposing that $P(e)$ holds more generally, and testing this hypothesis against further examples. Thus a common pattern is that abstraction produces the predicate $P$, while conjecture asserts its broader validity. 

\subsubsection{Refactoring} 

Once we have found suitable abstractions, there is the problem of refactoring. Suppose we have A used to define B used in C, now we change the definitions of A (new foundations) in a way which preserves B. In the formalism as given we need to reconstruct the whole hypergraph following B. Can we do better?  A  way to do it using the tools of computer science would be to have an abstraction barrier with named functions, whose referents can change.  Another possibility would be to make the original definition of B explicitly depend on A -- on the one hand every time we use B we have to refer to A, but then if we change A to A$'$ we can do it easily.  This is also related to ``concepts'' considered as isomorphic representations (such as abstract groups defined by multiplication law vs defined by homomorphisms, quotients and so on).

\subsection{AI models of mathematical discovery}
\label{ss:agentic}

We now illustrate how several concrete automated mathematical discovery systems instantiate the generic agentic loop of \S\ref{ss:models}.  Our purpose is not a full survey, but to show that the primitives of our framework (a growing knowledge hypergraph $\C_t \subset \St$, actions that propose/solve/compress, and learning signals from success and failure) correspond closely to mechanisms used in working AI systems. See also \cite{ColtonBundyWalsh2000Interestingness} for a survey of earlier work.

\paragraph{Deductive vs.\ inductive.}
\iffalse
Autonomous mathematical discovery (AMD) systems fall into two broad classes, 
deductive and inductive.  A deductive system makes conjectures expressed in formal logic,
tries to prove them, adds the proven theorems to its knowledge, and learns from both successful and failed attempts. Two recent deductive systems are Minimo \cite{poesia_learning_2024} and Fermat \cite{tsoukalas_learning_2025}. So far they have discovered and proved theorems in arithmetic and other elementary topics, but they could in principle be applied to large areas of mathematics.
\fi

Autonomous mathematical discovery (AMD) systems fall into two broad classes, 
deductive and inductive. A deductive system makes conjectures expressed in formal logic, 
tries to prove them, adds the proven theorems to its knowledge, and learns from both successful and failed attempts. Pioneering early work in this area, often termed automated theory exploration, includes systems like IsaCoSy \cite{johansson2009isacosy}, which systematically synthesized well-typed conjectures and attempted to prove them using the Isabelle/HOL deductive theorem prover. Foundational automated deduction engines like McCune's OTTER also demonstrated early discovery capabilities by autonomously finding novel single-axiom bases for algebraic structures \cite{mccune1993single}. Building on these symbolic foundations, two recent deductive systems leveraging modern machine learning are Minimo \cite{poesia_learning_2024} and Fermat \cite{tsoukalas_learning_2025}. So far they have discovered and proved theorems in arithmetic and other elementary topics, but they could in principle be applied to large areas of mathematics.

Inductive systems work with a database of instances of some mathematical object, find predicates which are true for all of the observed instances, and offer a subset of them as conjectures, chosen using heuristics which favor simplicity and likelihood of generalization.  These systems are more diverse and generally work within more specialized logical or programming frameworks.
An early inductive system is Graffiti \cite{fajtlowicz1988conjectures}, and its recent descendants TxGraffiti \cite{davila2024automated} and Graffiti$^3$. They are well known for conjectures in graph theory, formulated as inequalities on a predetermined set of graph invariants, which led to many published papers \cite{delavina2005history}. The most recent version (Graffiti$^3$) also adds new graphs chosen to try to falsify a conjecture \cite{davila2026graffiti3} (see also the influential \cite{wagner_constructions_2021}).
Many other early systems such as HR are discussed in the review \cite{colton2000automatic}.

A task closely related to inductive AMD is inductive program synthesis (IPS): given a list of (input,output) pairs, write a program to fit the functional relation which generalizes to new pairs. As an example, the input "1 2 3" and output "2 4 6" could be fit by a ``doubling'' function, ``(map ($\lambda$ x, * 2 x))''. IPS and AMD share many features -- compositionality, the need for search to find solutions, and especially the usefulness of finding and learning abstractions which capture broad patterns from previous examples. Recent examples are Dreamcoder and Lilo/Stitch \cite{ellis_dreamcoder_2020,grand_lilo_2023} which implement systematic and powerful methods for finding abstractions.

A recent and powerful extension of this task paradigm is {constructive algorithmic discovery} (CAD). Rather than synthesizing programs to fit a static list of predetermined examples, these systems synthesize executable code to actively construct extremal mathematical objects or discover novel algorithms evaluated against an open-ended mathematical scoring function. For instance, the FunSearch and AlphaEvolve systems \cite{romera2024mathematical,novikov2025alphaevolve} demonstrate how LLMs can be used to iteratively evolve programs that discover new, publishable bounds in extremal combinatorics.

We now give brief explanations of the various concepts and components of these AMD and IPS systems, organized as in \S \ref{ss:models}. First, each system is based on a mathematical or programming framework.  Minimo uses a simplified dependent type theory for Peano arithmetic (appropriately called Peano).  Fermat uses first order logic with axioms for arithmetic as implemented by the Z3 SMT solver.  Dreamcoder and Lilo/Stitch use a custom programming language
based on lambda calculus (so, similar to Lisp).  Graffiti relies on hand-coded definitions of invariants. Within these frameworks, all of the systems follow the four step iterative process of \S \ref{ss:models}.

Step 1 of the iteration is to make a conjecture (Minino and Fermat) or choose an IPS problem to solve (Dreamcoder/Lilo).\footnote{For Graffiti, choices of objects and invariants to consider are made by the user. } These are generated by modifying the existing corpus of proven
statements or solved IPS problems, by various methods.  Dreamcoder used a probabilistic generative model which combines previously learned functions to produce a new function, and then generates input-output pairs using the new function. Fermat combines proven theorems  using rules such as ``fix a variable to a constant value'' or ``compose two functions.'' Lilo uses a language model (LM) trained on internet data, while Minimo uses a LM trained only on its own corpus.

All of the models face the problem of selecting conjectures and problems at the right level of difficulty and likely to produce interesting results. In early systems this was done by heuristics; in modern systems it is done by passing information between the generator and prover models.  This is particularly systematic in Minimo, which obtains a ``difficulty'' signal from the prover (explained shortly) and uses it  to train the generator. Fermat evolves a model to judge interestingness of conjectures, employing the CAD paradigm.

Step 2 is to prove or solve. This is generally done by searching a vast combinatorial space for candidate solutions. Minimo uses Monte Carlo Tree Search (MCTS), a procedure which samples a large number of ``rollouts'' (successive actions chosen according to the policy) and keeps those with the largest reward. MCTS provides a probability for the rollout, which is used to derive the difficulty signal: an interesting problem is one which is just on the threshold of solvability. Dreamcoder and Lilo, on the other hand, solve IPS tasks using guided enumeration, where a neural recognition model directs a top-down search through the space of possible programs. Fermat does not have its own internal solver and instead outsources the proof search to an existing SMT solver, Z3 \cite{de2008z3}. Graffiti did not use a deductive prover at all; rather, it tested its conjectures empirically by checking them against a database of graphs to look for counterexamples.

Step 3 is to learn from the results of step 2. Success at proving or solving gives a training signal which can be used to improve the prover model by reinforcement learning. Minimo also uses ``hindsight experience replay'' -- a proof which does not succeed can be regarded as having succeeded at proving the intermediate results. Learning from the results involves not just getting better at proving, but also in developing abstractions. Dreamcoder and Lilo generate abstractions by finding common subgraphs, which can then be compressed by defining a new abstraction. 

Step 4 involves updating the knowledge corpus summarized in the hypergraph $\mathcal{C}_t$. One can define a utility function to assign a score to abstractions (as in Eq. 8 of \cite{bowers_top-down_2023}):
\begin{equation}\label{eq:stitchcost}
    U_{{\mathcal P},{\mathcal R}}(A) = \mbox{cost}({\mathcal P}) - \mbox{cost}(A) 
    - \mbox{cost}({\boldclass\mbox{Rewrite}}_{\mathcal R}({\mathcal P},A))
\end{equation}
where ${\mathcal P}$ is a set of definitions, $A$ is an abstraction
and ${\mathcal R}$ is a rewrite strategy (a specific way to use abstractions).
The cost of a statement (their Eq. (9)) is defined inductively (much like our complexity). In words, this is the gain achieved from rewriting the corpus $\mathcal P$ using $A$, minus the cost of $A$. \cite{bowers_top-down_2023} gives an efficient algorithm for proposing abstractions which minimize Eq. (\ref{eq:stitchcost}). The abstractions which are kept are those that maximize the utility function. It is particularly interesting to contemplate Eq. \ref{eq:stitchcost} and variants using the complexity measures proposed in Sec. \ref{sec:universal_importance} as universal objective measures for the importance of a theorem in mathematics. 

Fermat has as state S (a knowledge graph whose nodes come from our $\St$ and with an edge if $a$ was used by the action which generated $b$), actions A consisting of ``introduce conjecture," ``introduce definition," ``proof search") and rewards based on reproducing a given set of goals.
There is a nice set of conjecture generating heuristics. Interestingness is the intrinsic reward function evaluating an entity $s\in\St$ (our notation) in a given state. It is developed through evolutionary search.

What can we take from these systems?  Most of them started out {\it ab initio} with very little mathematical knowledge,\footnote{The exception is Lilo which made use of the OpenAI Codex LLM.} and all of them discovered significant concepts.  We regard them as proof of principle that automated mathematical discovery is possible. Their common adherence to the paradigm summarized in \S \ref{ss:models} could also be taken as evidence that the ingredients of the paradigm -- in particular conjecturing and compression -- are important.

\subsection{Tests of mathematical ability}
\label{ss:tests}

Benchmarks are crucial to measuring AI progress, yet have quickly been getting saturated in recent years. 
One interesting class of synthetic benchmarks that has been less exploited is to choose some large class of tasks with a simple systematic definition, and ask that the system's performance on the tasks improves with time. For example, a number theory system might be asked to solve randomly sampled Diophantine equations or show that they cannot be solved.  As a function of the degree and number of variables of the equations, this rapidly becomes very difficult.  But there are many general patterns and methods, and a AMD system given this task could much enhance its performance by discovering them and proving theorems about their correectness and applicability.  The task does not change, but the range and sophistication of the system's solutions can grow.

While correctness of the solutions is a test, a much stronger test -- which we would consider a defining aspect of doing mathematics -- is
to prove or at least give strong arguments for the solutions.  Similarly, if a theorem is proven to help in this task, the system should have some argument that it does help.
These reasons could be human-style explanations, but we do not insist on this. The reasons could also have  varying degrees of merit. As an example, the system might estimate the value of each new theorem for use in proving other theorems, and rank its proposals by this ``utility'' measure.  This counts as a reason to the extent that its measure is well founded and improves proving performance.  As other examples, early work on AMD such as Graffiti and HR \cite{fajtlowicz1988conjectures,davila2024automated,colton2000automatic}
had hand-coded heuristics for judging interestingness; these would count, but 
a system such as Fermat \cite{tsoukalas_learning_2025} which learns an interestingness function would do better on this criterion.

The final test of mathematical discovery, used also to evaluate human mathematicians, is (1) solving an open problem, for example discovering a proof of a famous conjecture, or (2) discovering a concept (abstraction, definition, structure, mathematical object). (2) A discovered concept should either aid in (1), or it should open up many new directions and thus be used often going forward. We can envision applying both of these discovery benchmarks to AI systems. 

We can envision an AI mathematical society (a multi-agent AI system), where they pose their own problems, find proofs, document (and perhaps rank) useful concepts and abstractions, and so on. One can run such an AI society autonomously, and study its progress; it would be interesting to develop quantitative ways to score this progress. In \S \ref{ss:criteria} we discuss some criteria to evaluate automated mathematical discovery. 

We can also measure how much an autonomously running AI system makes discoveries that aid or are of interest to humans, which is ultimately what we care about. Lists of human-specified open problems are valuable, and we are already seeing this in the Erdös problems \cite{BloomErdosProblems} and FrontierMath's recent OpenProblems list \cite{EpochAI_FrontierMath_OpenProblems}. We need measures of how much an AI-discovered concept is used by humans in the future, which can be done through specific types of citation-tracking.   
\section{Thoughts on human mathematics}
\label{s:human}

In light of all this, what can we say about human mathematics? In this section, we tackle a few natural questions that arise from the preceding discussions and mention some caveats. 

\subsection{Can we define an ``ideal’' human mathematics $\overline{\HH}$ ? }

To begin with, how do we define human mathematics? The obvious definition is to look at what humans know at time $t$, formalize it in some way and let this be $\HH_t$. But, humans do have the intuition that human mathematics is universal or at least strives towards the universal. It is interesting to contemplate whether this has some limiting form -- to what extent is there a notion of an ``ideal" human mathematics? Before listing three possible definitions, there is a caveat which we have suppressed until now. $\HH$ is better understood as a measure on $\UU$ rather than a subset, we should not really expect it to have a sharp edge. The "interesting number paradox" is at work here. First discussed by E. Beckenbach \cite{beckenbach1945interesting} and popularized by M. Gardner, if each number is  either interesting or not, then the \textit{smallest} uninteresting number surely is of interest. Consequently, interest must fade away gradually. Similarly, for finding the edge of $\HH$ . Self-reference surely builds new statements in $\HH$ in the same way the  metamathematics of $\UU$, formalized as logic, itself, becomes a part of $\UU$. But suppressing such subtleties, here are some possibilities:
\begin{enumerate}
    \item 
The colimit of $\HH_t$ granting that humans exist and keep doing math for arbitrarily long $t$.
Not precise, but substitute an AMD agent for human and it becomes precise.
\item By human judgment, in other words we postulate an oracle which has access to all of $\UU$ and asks the humans ``does statement $s$ belong in $\overline{\HH}$ ?’'
\item The other way around - if we had an oracle, what questions would we ask it?   These questions are the backbone of the extrapolation of $\overline{\HH}$.

If $\overline{\HH}$ makes sense, how unique is it?
\end{enumerate}

The oracle is particularly relevant in the context of AI. It suggests a variety of thought experiments. Suppose we had two groups of mathematicians each granted access to the oracle but unable to communicate with each other.  They would quickly learn a lot, i.e. their respective $\HH_t$ would grow rapidly.  But would their $\HH_t$’s be similar or would they wander off in very different directions indefinitely?  How rapidly would they grow? This latter question might require a computationally constrained oracle to be interesting.

\subsection{AI in the oracle limit}

Given the current trajectory of AI systems, it is natural to wonder about the fate of mathematics in the limit where AI systems have become arbitrarily powerful compared to humans.  It may be worth emphasizing that humans will not stop doing mathematics, although the way it is done will change dramatically. Rather, human mathematics will likely flourish in this limit, and likely be more popular and accessible than ever. 

In the future we will use our mathematical and physical heritage to understand the vast, infinite structures of the Platonic worlds, as they are reported to us. These worlds have both practical and aesthetic value. In this limit, the AI oracle will help speed up the work of human mathematicians, and will uncover new directions for them to ponder. But ultimately, the bandwidth of the human mind serves as a bottleneck, since human mathematicians will always want to understand. 

Even the concept "to understand" will require conservation. Just as the GPS on your phone may have damaged your sense of direction in a cityscape, so reliance on AIs to handle "details" may blur what you understand. Feynman: "That which I cannot reproduce I do not understand." 

Since AIs will always be computationally bounded, the combinatorial explosion of mathematics means that no matter how many autonomous discovery agents we launch, there will always be plenty of directions for humans to explore as well (likely with the aid of AI systems) which autonomous discovery agents almost surely will have left untouched. We will need to learn the most fruitful interaction modes with our agents.

\subsection{How does the human mind traverse the Platonic worlds?}

\subsubsection{Wishful Thinking}

Wishful thinking has always been an under-rated tool for mathematical progress. We hope a statement is true, admire its aesthetic qualities, and try to prove it true. This is less principled and possibly more important than \textbf{conjecture}, which is a precise claim based on structured evidence. We work both inductively from examples, and from the magnetic pull of abstraction. 
Once one has had a crazy idea work, they become addictive. As we discussed in  \S \ref{sec:conjecture} and \S \ref{sec:conj_vs_abs}, we may be able to systematize these for AI systems. AI's may observe dramatic connectivity changes in the hypergraph $\Hum$ if a certain candidate statement were to be true. An AI could search over candidate statements and abstractions for which known examples and constructions would become special cases, and then observe that many disparate ones would suddenly emanate from this new candidate node. Alternatively, the new candidate node would connect different clusters (fields of study) in $\Hum$. Humans have the habit of modifying their objective to see emergent structure. Mathematics is part show business, our results should tell a story, some story, to attract the attention of our friends. This suggests communication and persuasion as an important component in mathematical discovery, and we need to add it in multi-agent AI systems. AIs can succeed at this too and learn to\textbf{ wish} as well as \textbf{conjecture}. 

\subsubsection{Hierarchy and abstraction}

It is generally accepted that the capabilities exhibited by both machine and biological intelligences are enabled fundamentally by the hierarchical and compositional structure of natural data.\footnote{The role of this hierarchical and compositional structure is being actively studied by scientists in pursuit of an improved understanding of machine learning. See e.g. \cite{saxe2013learning,lin2017does,cagnetta2024deep, PoggioFraser2024CompositionalSparsity}.} This structure allows learning to evade the curse of dimensionality, since the learning agent presumably only needs to learn a manageable number of low-order correlations but in increasingly abstract representation spaces. 

For example, progress in science is possible because it is largely modular: Much can be studied in isolation and often crude approximation gives serviceable agreement with a richer reality. Famously, Galileo studied "falling" by rolling balls down ramps fretted with musical strings; his musical sense allowed him to adjust the frets until the beat was steady. Fortunately for science, angular inertia of the balls, their interaction with the frets, and their air resistance can be safely ignored. Earlier and later examples are the pendulum clock and quantum electrodynamics (QED).

In natural language processing, words can be grouped into part-of-speech tags (noun, verb, adjective, etc), which are then grouped into phrases (noun phrases, verb phrases, etc), and so on \cite{JurafskyMartin2009}. Linguists painstakingly hard-code these hierarchical structures in parse trees, which aids our understanding of natural language. Modern LLMs do not need this annotation; presumably they spontaneously discover these hierarchies (and much more) through gradient-based learning. Many linguistic conventions are quite strict but hard to enumerate (Consider adjective order; one would not say "a fat big cat".) but LLMs master these as well as the ones we recall learning in high school. In the natural course of their development they will learn much more about how both human and formal math work in practice than we could possibly tell them ourselves. Finding a useful intermediate proof objective feels like finding an inspired Go move. As with Go, we may help them best by knowing when to get out of the way.

In mathematics, the formal system is roughly analogous to the raw pixel or token level data; we think of it as "machine level."  But human mathematicians think in abstractions that may be 10 or more layers deep. We have formalized these layers by hand, the way linguists hand-annotate a corpus of text with parse trees. This human-constructed partial order of definition and lemma might illuminate the hierarchical nature of hypergraphs $\Hum^{(i)}$ at varying levels of abstraction, perhaps distinguishing it from the larger $\U^{(i)}$. The $\U^{(i)}$ may be full of weird redundancies (as in our "and" example), and simultaneously much less amenable  to  systematic nested compression in the form of definitions and lemmas. In the modern deep learning paradigm, we would expect that training AIs on mathematics, both formal and informal, respectively, should induce them to spontaneously reason in similar abstraction. 

\subsection{Non-universal importance measures}

Above we discussed universal objective measures of interestingness and importance, which might select out human mathematics. Of course there are many non-universal measures. For example, whether a given theorem is surprising, which is based on our knowledge at a specific moment in time. Interestingly, this can be measured quantitatively in AI systems, which are fundamentally probabilistic in nature, in terms of the probability that an AI model assigns to a new statement. See also \cite{BengioMalkin2024AItheory}. 

Our sociological dynamics clearly enters \cite{venkatesh2024some} as well. A proof that has evaded many attempts over decades or centuries  becomes important. 

Finally, clearly much of human mathematics has arisen in the service of advancing various sciences and is a consequence of our pursuit of understanding a physical (3+1)-dimensional world. As AI for mathematics progresses, we will likely learn more about the relationship between mathematics as a precise tool for science and mathematics as a universal structure that is being uncovered by our human minds. 

\section{Paths forward}
\label{s:paths}

\subsection{Computational Metamathematics}

Our AI mathematical agents will enrich our understanding of the foundations of mathematics in ways orthogonal to the work of logicians. The hypergraphs of mathematics call out for modeling; what are their statistical properties? What is their coarse geometry? Are there  rules of thumb that we and our agents can learn that will help us to discover what we will most want to know?

As an increasing fraction of human mathematics is formalized, we will soon have compiled a large segment of the graph $\Hum$ in various formal systems. This means that we can embark on computational analysis of the structure of $\Hum$ and analyze some of the questions posed above in terms of importance of theorems and definitions and the role of conjecture. Including dates of mathematical results also allow study of the evolution of $\Hum$ through the course of human history. Training AIs on the historical time-dependence of $\Hum$ should help extrapolate our joint progression into the future. 

Given the doubly exponential growth of $\U$, it is unclear to what extent its structure can be probed by brute force. One possibility is to let our AIs venture into a small neighborhood outside of $\Hum \subset \U$, which we may call $\Hum_\epsilon$. Studying $\Hum_\epsilon$, the statistical properties of lengths (and compressiblity) of statements in $\Hum_\epsilon$, may provide us with additional insight into $\Hum$ and $\U$ and the degree to which $\Hum$ looks special as a subset of $\U$. This can be iterated: Explore, collapse to an $\Hum$-like core, explore more, collapse again to a larger $\Hum$-like core, etc. This may be related to how humans do mathematics, as the introduction of a concept and a definition is often what opens up a new vista. 

We outlined two broad ways to understand what determines the subset of mathematical facts $\HH \subset \UU$ which humans find worthwhile. One is a global and universal picture in which intrinsic properties of $\UU$ itself considered as a mathematical object can determine this,
for example $\HH$ might optimize some objective local observable. The other is by modeling the process of exploring and creating mathematics, leading to simulated histories of mathematical development which will combine both the contingent and the universal.  Work on automated mathematical discovery already shows ways that this can be done and we can expect significant progress in the near term.  It will be fascinating to find out whether such AI systems work in similar ways to humans and produce comparable mathematics, or whether there are alien forms of mathematics yet to be developed.

Both approaches are clearly interesting and need not be in conflict.  For example, an agent model is usually designed to optimize a reward function -- here measuring the quality of the mathematics it produces.  While such a computable reward function may look like a poor and noisy
approximation to the global quality measure postulated in the universal picture, nevertheless if the signal to noise ratio is high enough then the two could have the same optima.

Suppose we are disappointed and the Platonic hypergraphs in $\UU$  do not contain particularly notable structures, and it proves difficult to algorithmically pick out $\HH \subset \UU$. Then what is mathematics? Perhaps mathematics is about distilling concepts that our human minds have found interesting and useful over time, gated by biological evolution and interaction in the physical world. Then mathematics is more like art and/or engineering -- it is invented or designed to meet aesthetic and/or functional criteria. If this is the case, then for AIs to discover mathematics, they must first understand us, and to do so fully, be raised in our rich 3+1-dimensional physical world. This falls into the general category of intrinsic vs extrinsic motivation; we can give  the AIs problems derived from our experience in the physical world as extrinsic motivation. 

If $\mathcal{U}$ is ever to be explored empirically, quantum computers might be crucial since they access exponential orthogonal storage and double-exponential "nearly-orthogonal"  storage. Of course, in quantum computing only patterned wave functions can be usefully readout (as through Fourier transform) but patterns in $\mathcal{U}$ are precisely what we would look for. 

\subsection{Criteria for autonomous mathematical discovery}
\label{ss:criteria}

At present there is a paradox in AI.  On the one hand there is amazing sustained progress leading many to predict that AI's will match top human performance at many, possibly all, mathematical tasks within a decade. On the other hand, as yet there is no generally accepted case of an AI making a mathematical discovery on its own\footnote{In an amusing analogy in 2004 skeptics of Perelman's proof of the Poincare conjecture often pointed out that "no topology theorem had ever received its first proof via differential equations". Sometimes things change dramatically.} beyond highly constrained domains where the discovery process is a computational search through a predetermined space.\footnote{Examples: McCune's OTTER system discovered a single axiom which defines an abstract group, Google DeepMind's AlphaEvolve and FunSearch improved bounds in extremal combinatorics and discovered interesting counterexamples.}
While there are many joint human-AI discoveries (e.g. see the conjectures of Graffiti), these involved human supervision and a skeptic can maintain the position that they were all made by humans using AI as a tool. Clearly this paradox cannot continue; to justify the optimists AIs must start making their own discoveries soon.

\begin{table}[t]
\centering

{\bfseries AMD Criteria: Combined Ratings from Frontier Models\par}
\vspace{4pt}
\vspace{2pt}
\renewcommand{\arraystretch}{1.4}
\setlength{\tabcolsep}{5pt}
\footnotesize
\begin{tabular}{>{\raggedright\arraybackslash}p{2.8cm}|c|c|c|c|c}
\toprule
\rowcolor{darkblue}
\textcolor{white}{\textbf{Criterion}} &
\textcolor{white}{\textbf{Minimo}} &
\textcolor{white}{\textbf{Lilo\,/\,Stitch}} &
\textcolor{white}{\textbf{Fermat}} &
\textcolor{white}{\textbf{Graffiti}} &
\textcolor{white}{\textbf{AlphaEvolve}} \\
\rowcolor{medblue}
\textcolor{white}{\scriptsize\textsf{}} &
\textcolor{white}{\scriptsize\textsf{Cl\,/\,Ge\,/\,GPT}} &
\textcolor{white}{\scriptsize\textsf{Cl\,/\,Ge\,/\,GPT}} &
\textcolor{white}{\scriptsize\textsf{Cl\,/\,Ge\,/\,GPT}} &
\textcolor{white}{\scriptsize\textsf{Cl\,/\,Ge\,/\,GPT}} &
\textcolor{white}{\scriptsize\textsf{Cl\,/\,Ge\,/\,GPT}} \\
\midrule
%%
%% Columns: Minimo, Lilo/Stitch (merged), Fermat, Graffiti, AlphaEvolve
%%
%% Lilo/Stitch merged: take the more generous of the two per rater.
%%
%% Claude:  Lilo  P N P N Y P P P P Y   Stitch N N P N Y P P N P N  -> merged: P N P N Y P P P P Y
%% Gemini:  Lilo  N N N N Y Y Y N Y P   Stitch N N N N Y Y Y N Y N  -> merged: N N N N Y Y Y N Y P
%% GPT:     Lilo  N N N N P P P N P N   Stitch N N N N P P P N P N  -> merged: N N N N P P P N P N
%%
%% C1: Open-ended math language
\rowcolor{white}
C1: Open-ended language
  & \tri{\yes}{\yes}{\pmark}
  & \tri{\pmark}{\no}{\no}
  & \tri{\pmark}{\yes}{\pmark}
  & \tri{\pmark}{\no}{\no}
  & \tri{\pmark}{\no}{\pmark} \\
\rowcolor{lighterblue}
C2: Verifiable proofs
  & \tri{\yes}{\yes}{\yes}
  & \tri{\no}{\no}{\no}
  & \tri{\yes}{\yes}{\pmark}
  & \tri{\no}{\no}{\no}
  & \tri{\pmark}{\no}{\pmark} \\
%% C3: Novelty detection
\rowcolor{white}
C3: Novelty detection
  & \tri{\yes}{\yes}{\pmark}
  & \tri{\pmark}{\no}{\no}
  & \tri{\pmark}{\pmark}{\pmark}
  & \tri{\yes}{\no}{\pmark}
  & \tri{\yes}{\no}{\pmark} \\
%% C4: Proposes & proves theorems
\rowcolor{lighterblue}
C4: Proposes \& proves
  & \tri{\yes}{\yes}{\yes}
  & \tri{\no}{\no}{\no}
  & \tri{\yes}{\yes}{\pmark}
  & \tri{\pmark}{\pmark}{\no}
  & \tri{\pmark}{\pmark}{\pmark} \\
%% C5: New definitions/concepts
\rowcolor{white}
C5: New definitions
  & \tri{\no}{\no}{\no}
  & \tri{\yes}{\yes}{\pmark}
  & \tri{\yes}{\yes}{\yes}
  & \tri{\no}{\yes}{\no}
  & \tri{\no}{\yes}{\pmark} \\
%% C6: Selects interesting discoveries
\rowcolor{lighterblue}
C6: Selects discoveries
  & \tri{\yes}{\yes}{\no}
  & \tri{\pmark}{\yes}{\pmark}
  & \tri{\yes}{\yes}{\yes}
  & \tri{\yes}{\yes}{\yes}
  & \tri{\yes}{\yes}{\yes} \\
%% C7: Reasons for selection
\rowcolor{white}
C7: Reasons for selection
  & \tri{\pmark}{\no}{\no}
  & \tri{\pmark}{\yes}{\pmark}
  & \tri{\pmark}{\yes}{\yes}
  & \tri{\pmark}{\no}{\pmark}
  & \tri{\pmark}{\pmark}{\pmark} \\
%% C8: Research program
\rowcolor{lighterblue}
C8: Research program
  & \tri{\pmark}{\no}{\pmark}
  & \tri{\pmark}{\no}{\no}
  & \tri{\pmark}{\pmark}{\pmark}
  & \tri{\no}{\no}{\no}
  & \tri{\pmark}{\pmark}{\pmark} \\
%% C9: Independent validation
\rowcolor{white}
C9: Validation
  & \tri{\pmark}{\yes}{\pmark}
  & \tri{\pmark}{\yes}{\pmark}
  & \tri{\pmark}{\yes}{\yes}
  & \tri{\yes}{\yes}{\yes}
  & \tri{\yes}{\yes}{\yes} \\
%% C10: Closed-loop expansion
\rowcolor{lighterblue}
C10: Closed-loop expansion
  & \tri{\pmark}{\yes}{\pmark}
  & \tri{\yes}{\yes}{\no}
  & \tri{\pmark}{\yes}{\pmark}
  & \tri{\no}{\no}{\no}
  & \tri{\pmark}{\yes}{\no} \\
\bottomrule
\end{tabular}
\caption{\label{table:amd_ratings} Ratings of various AMD systems according to our AMD criteria in Fig. \ref{fig:criteria}, according to Claude Opus 4.6 (Cl), Gemini 3.1 Pro (Ge), and ChatGPT 5.4 Pro (GPT). 
\yes\ = satisfied,\quad
\pmark\ = partial,\quad
\no\ = not satisfied. The models also provided explanations for each rating, which we have not included here.}
\end{table}

There are several distinct notions of autonomous mathematical discovery. In one case, the AI autonomously makes its own discoveries which it finds interesting and important. This leads to the development of a separate strand of "AI mathematics," which may be already known or uninteresting to humans. The other case, which the world eagerly awaits, is for AIs to autonomously make discoveries that \it humans \rm find interesting or important. 

These possibilities also differ in how much prior mathematics the system is given. A system given access to all of mathematics must go beyond all that is known.\footnote{This is time-dependent, making it interesting to train an LLM only on mathematics up to time $t$ and compare its discoveries with those of humans. } At the opposite extreme one could envision a ``MathZero'' system \cite{mcallester_mathzero_2020} which, by analogy with AlphaZero, is given only the foundations and is tasked with rediscovering all of human mathematics. All of the above count as different forms of autonomous discovery. 

In Fig. \ref{fig:criteria} we list ten criteria in an attempt to cover both extremes and the spectrum in between. A passing score, by any agent, would signal the kind of remarkable advance that we have come to expect from this field. In Table \ref{table:amd_ratings}, we have various state of the art AI systems (ChatGPT 5.4 Pro, Claude Opus 4.6, and Gemini 3.1 Pro), grade the AMD systems we discussed in \S \ref{ss:agentic} according to the criteria in our Fig. \ref{fig:criteria}. It is interesting to note that the most recent system, Fermat, also scores the highest on all criteria, demonstrating clear progress for AMD systems. 

\begin{figure}[h!]
  \centering
  \includegraphics[width=0.9\linewidth]{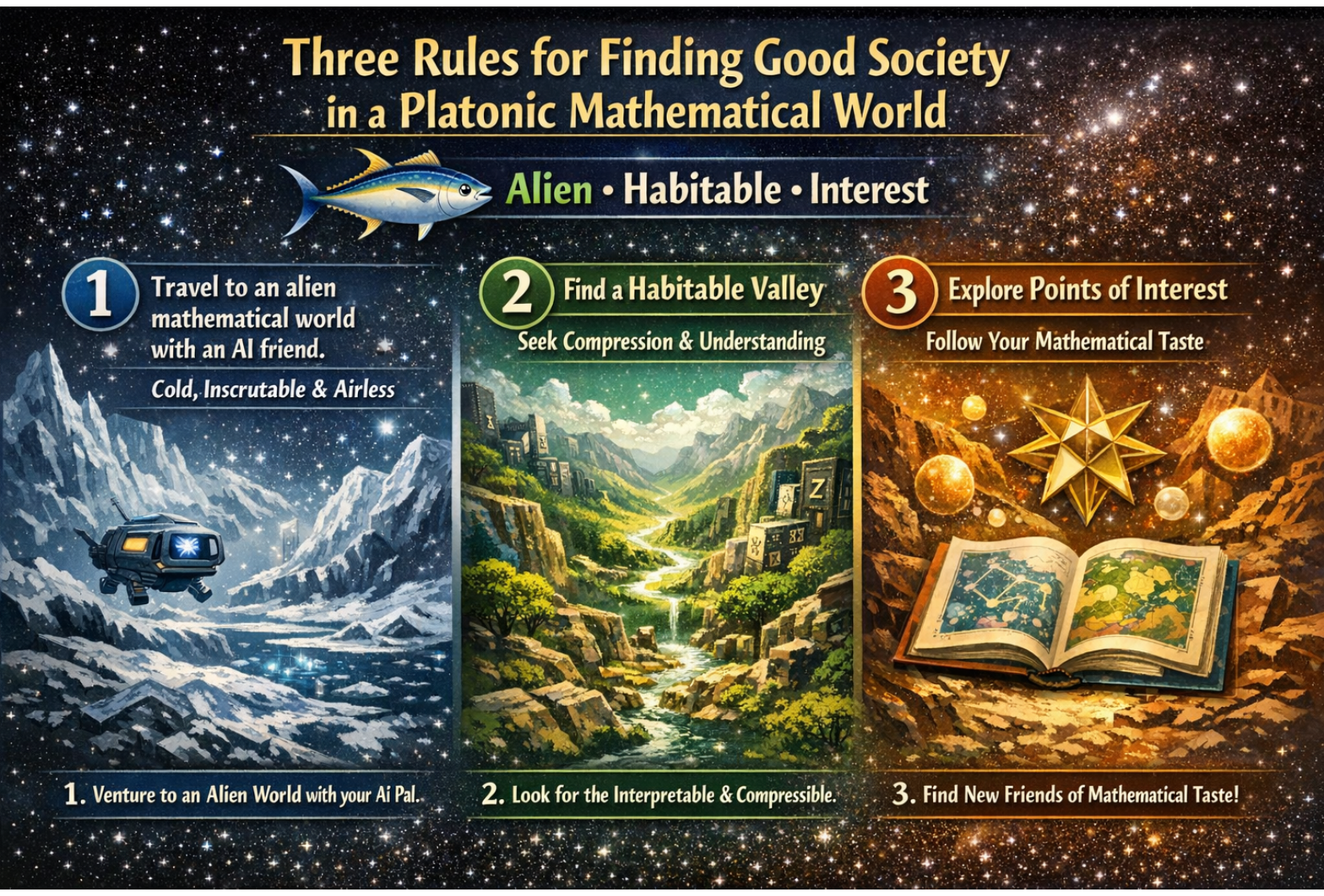}
  \caption{Figure generated by ChatGPT}
  \label{fig:AHI}
\end{figure}

These are criteria for autonomous AI discovery, but of course we are not creating AIs just to leave all the fun to them! The point is to create new collaborators, companions and friends with which to explore. To this end, let us give
\textbf{Three Rules  for Finding Good Society in a Platonic Mathematical World}, Fig. \ref{fig:AHI}.

\section{Acknowledgments}

MB and MRD thank the Simons Collaboration on Physics of Learning and Neural Computation (SFI-MPS-POL-00012574-09). MRD thanks David McAllester and Gerald Jay Sussman for long-running discussions on these topics.

\appendix

\section{Examples of hypergraphs}
\label{s:exhyper}

This section can be compared with chapter 1 of the HoTT book \cite{aczel2013homotopy}.
We give hypergraph representations of a few definitions and propositions proven there.

We recall that in Peano arithmetic, the natural numbers are defined to be either $0$ or the successor $Sn$
of a natural number $n$.  For us $0\in \St_0$ and $S$ is a $(1,1)$-hyperedge.
Addition is defined recursively, using  the $\mathbb{N}$-Elimination hyperedge, denoted as \texttt{Rec}.

\subsection{The \texttt{Rec} Hyperedge}
The \texttt{Rec} (4,1)-hyperedge represents the principle of dependent mathematical induction implemented as a computational rule.   Its inputs are
\begin{enumerate}
    \item \textbf{Motive ($P$):} A type family $P : \mathbb{N} \to \textbf{Univ}$ that determines the type of the result for each input. This allows \texttt{Rec} to handle both simple recursion (where $P$ is constant) and induction (where $P$ varies).
    \item \textbf{Base Case ($z$):} The value to return if the target is $0$. It must satisfy the typing $z : P(0)$.
    \item \textbf{Step Function ($s$):} A function defining the inductive step. It must have the type $\Pi(k:\mathbb{N}).(P(k) \to P(S(k)))$, accepting the current index $k$ and the recursive result for $k$ to produce the result for $S(k)$.
    \item \textbf{Target ($n$):} The natural number being analyzed (decomposed), which drives the reduction.
\end{enumerate}
\noindent The output of this hyperedge is the term $\texttt{rec}(P, z, s, n)$, which automatically possesses the typing edge to $P(n)$.

\subsection{The Definition of Addition}

We define addition, denoted $add(m, n)$ or synonymously $m+n$,
by applying the \texttt{Rec} hyperedge to the target $n$. To satisfy the formal requirement of Dependent Type Theory, we must provide all four inputs: the Motive, the Base Case, the Step Function, and the Target.

\subsubsection{Input Configuration for $m+n$}

\begin{enumerate}
    \item \textbf{Input 1: The Motive ($P$).} 
    The motive determines the type of the result based on the input. For addition, the result is always a natural number regardless of $n$. Thus, we supply the \textbf{Constant Family}:
    \[ P :\equiv \lambda (\_ : \mathbb{N}). \mathbb{N} \]
    
    \item \textbf{Input 2: The Base Case ($z$).}
    This input must have type $P(0)$, which reduces to $\mathbb{N}$. Since $m + 0 = m$, we connect the variable node $m$:
    \[ z :\equiv m \]

    \item \textbf{Input 3: The Inductive Step ($s$).}
    This input must have type $\Pi(k:\mathbb{N}).(P(k) \to P(S(k)))$. Under our constant motive, this simplifies to $\mathbb{N} \to \mathbb{N} \to \mathbb{N}$. The function receives the index $k$ and the accumulated value $v$ (where $v = m+k$) and returns the successor:
    \[ s :\equiv \lambda k. \lambda v. S(v) \]

    \item \textbf{Input 4: The Target ($n$).}
    The number on which we recurse.
\end{enumerate}

\subsubsection{Resulting Node and Typing}
The output node of this hyperedge is the term $\texttt{rec}(P, m, s, n)$.
The \texttt{Rec} rule inherently generates a typing edge from this result to the type determined by applying the motive to the target:
\[ \text{Result} \to P(n) \xrightarrow{\beta} \mathbb{N} \]
Thus, the operation preserves the static typing required by Peano Arithmetic.

\begin{tikzpicture}[scale=0.8, transform shape,
    node distance=1.5cm and 2.5cm,
    term/.style={circle, draw=black, very thick, fill=white, minimum size=0.9cm, inner sep=2pt, font=\bfseries},
    ctor/.style={circle, draw=blue!80!black, very thick, fill=blue!10, minimum size=0.9cm, font=\bfseries},
    op/.style={rectangle, draw=red!70!black, fill=red!10, thick, inner sep=5pt, font=\small\ttfamily},
    link/.style={-Latex, thick, draw=gray!80},
    scope/.style={draw=green!50!black, dashed, thick, fill=green!5, rounded corners},
    motive/.style={rectangle, draw=purple!80!black, thick, fill=purple!10, rounded corners}
]

% --- Inputs ---
\node[term, label=left:$m$] (m) {};
\node[term, below=2.5cm of m, label=left:$n$] (n) {};

% --- Input 1: The Motive (Constant Family) ---
\node[motive, right=1cm of m, label=above:\textcolor{purple!60!black}{1. Motive}] (P) {$\lambda \_.\mathbb{N}$};

% --- Input 3: The Step Function ---
% Subgraph for \lambda k. \lambda v. S(v)
\node[term, right=3cm of P, label=below:$v$] (v) {};     
\node[term, above=0.5cm of v, label=left:$k$] (k) {};      
\node[ctor, right=1cm of v] (Sv) {$S$};                    
\draw[link] (v) -- (Sv);

\begin{scope}[on background layer]
    \node[scope, fit=(k) (v) (Sv), label=below:\textcolor{green!40!black}{\textit{Step Scope}}] (step_box) {};
\end{scope}

\node[op, right=0.5cm of step_box, label=above:3. Step] (lam_step) {$\lambda k.\lambda v$};
\draw[link] (Sv) -- (lam_step);

% --- The Rec Hyperedge (4 Inputs) ---
\node[op, below=1.5cm of lam_step, align=center, minimum width=2.5cm] (rec) {\textbf{Rec}$_\mathbb{N}$};

% Wiring the 4 Inputs
% 1. Motive
\draw[link, purple] (P) to[out=-45, in=135] (rec.north west);

% 2. Base Case (m) - connects to 'm'
\draw[link] (m) to[out=0, in=180] (rec.west);
\node[above left=0.1cm of rec.west, font=\scriptsize] {2. Base};

% 3. Step Case (The function)
\draw[link] (lam_step) -- (rec);

% 4. Target (n)
\draw[link] (n) to[out=0, in=225] (rec.south west);
\node[below left=0.1cm of rec.south west, font=\scriptsize] {4. Target};

% --- Result ---
\node[term, right=1.5cm of rec, label=right:Result] (res) {};
\node[below=0.1cm of res, color=gray] {$m+n$};

\end{tikzpicture}

\subsubsection{Computational interpretation}

The hypergraph above represents the definition of addition in a static way.
Suppose we apply it to specific numbers such as $2+2$; the result $4$ will be
the output of a larger ``computational'' or ``dynamic'' hypergraph pictured below.
It is created by adding a new type of $(1,1)$-hyperedge, sometimes called ``iota
reduction.''  This edge takes the output of a structural hyperedge and carries out
one step of evaluation.   Consider the example of $add(2,2)$; this evaluates to the
term 
$$\mathbf{Rec}(\mathbb{N}, \mathbf{2}, \text{Step}, \mathbf{2}).$$

This is already a node of $\St$, but to get a node representing a
simplification of it we would follow an iota reduction
edge whose output is determined by the definitions of its inputs (the subterms) and is
$$\text{App}(\text{plus}, 2, 2) \xrightarrow{\textbf{Iota-Red}} S(\text{App}(\text{plus}, 2, 1))
$$
Continuing along further Iota-Red
edges would eventually produce $SSSS0$, the unary representation of $4$.

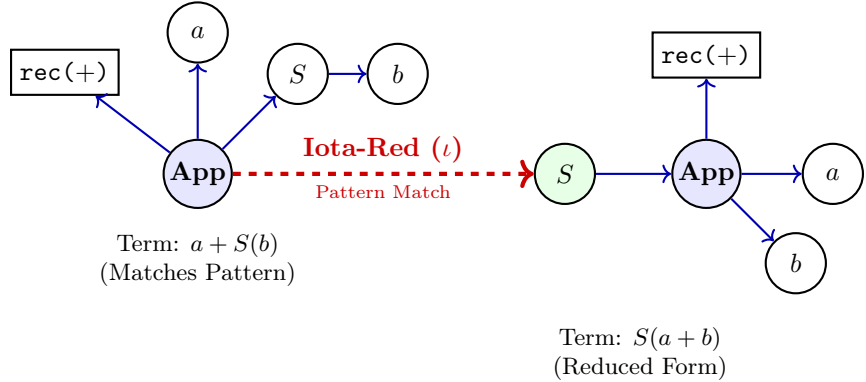
\begin{figure}[ht]
    \centering
    \begin{tikzpicture}[
        node distance=1.5cm and 2cm,
        % Styles for different node types
        term/.style={circle, draw=black, thick, minimum size=8mm, inner sep=1pt, font=\bfseries},
        const/.style={rectangle, draw=black, thick, minimum size=6mm, font=\ttfamily},
        % Styles for edges
        construct/.style={->, blue!70!black, thick},   % Blue for building terms
        compute/.style={->, red!80!black, ultra thick, dashed}, % Red for reduction
        label_text/.style={font=\small\sffamily, color=gray}
    ]

    % --- LEFT SIDE: The Expression a + S(b) ---
    
    % The Application Node (The "Redex")
    \node[term, fill=blue!10] (App) {App};
    
    % Inputs to App
    \node[const, above left=1cm of App] (Plus) {rec($+$)};
    \node[term, above=1cm of App] (VarA) {$a$};
    \node[term, above right=1cm of App] (S_node) {$S$};
    \node[term, right=0.5cm of S_node] (VarB) {$b$};
    
    % Construction edges
    \draw[construct] (S_node) -- (VarB); % S points to b
    \draw[construct] (App) -- (Plus);
    \draw[construct] (App) -- (VarA);
    \draw[construct] (App) -- (S_node);
    
    % Label for Left Side
    \node[below=0.2cm of App, align=center, font=\small] {Term: $a + S(b)$ \\ (Matches Pattern)};

    % --- RIGHT SIDE: The Reduction S(a + b) ---
    
    % The Outer Constructor
    \node[term, right=4cm of App, fill=green!10] (Result_S) {$S$};
    
    % The Inner Recursion
    \node[term, right=1cm of Result_S, fill=blue!10] (Inner_App) {App};
    
    % Inputs to Inner App
    \node[const, above=0.8cm of Inner_App] (Plus_Inner) {rec($+$)};
    \node[term, right=0.8cm of Inner_App] (VarA_Inner) {$a$};
    \node[term, below right=0.8cm of Inner_App] (VarB_Inner) {$b$};
    
    % Construction edges for Result
    \draw[construct] (Result_S) -- (Inner_App);
    \draw[construct] (Inner_App) -- (Plus_Inner);
    \draw[construct] (Inner_App) -- (VarA_Inner);
    \draw[construct] (Inner_App) -- (VarB_Inner);
    
    % Label for Right Side
    \node[below=1.5cm of Result_S, xshift=1cm, align=center, font=\small] {Term: $S(a+b)$ \\ (Reduced Form)};

    % --- THE IOTA EDGE ---
    
    \draw[compute] (App) -- node[above, font=\bfseries\color{red}] {Iota-Red ($\iota$)} 
                            node[below, font=\scriptsize\color{red}] {Pattern Match} (Result_S);

    \end{tikzpicture}
    \caption{\textbf{The \texttt{Iota-Red} Hyperedge.} 
    The blue solid lines represent static construction (Application). 
    The red dashed line represents the computational step. 
    Because the second argument matches the constructor $S$, the graph explicitly links the expression $a+S(b)$ to its reduction $S(a+b)$.}
    \label{fig:iota_red}
\end{figure}

\subsection{Example: The Doubling Function}
Following \S 1.9 of the HoTT Book, we define the operation $\texttt{double}(n) = 2n$ using the recursion principle for natural numbers. The definition in standard syntax is:
\begin{align*}
    \texttt{double}(0) &:\equiv 0 \\
    \texttt{double}(S(n)) &:\equiv S(S(\texttt{double}(n)))
\end{align*}

In the hypergraph formalism, this function is constructed by instantiating the \texttt{Rec} hyperedge with specific inputs:
\begin{enumerate}
    \item \textbf{Motive (Return Type):} The constant family $\mathbb{N}$.
    \item \textbf{Base Case:} The node $0$ (mapping input $0$ to result $0$).
    \item \textbf{Step Case:} A function $g : \mathbb{N} \to \mathbb{N} \to \mathbb{N}$.
\end{enumerate}

\paragraph{The Step Function Subgraph.}
The step function $g$ represents the logic: "Given the previous number $n$ and the computed result of the recursion $y$ (where $y = \texttt{double}(n)$), return $S(S(y))$."
\begin{itemize}
    \item We generate a dependency subgraph $\mathcal{D}(n, y)$ containing variables $n$ and $y$.
    \item We apply the constructor $S$ to $y$ twice, creating the chain $y \to S(y) \to S(S(y))$.
    \item Crucially, the variable $n$ is unused (the operation is purely algebraic on the result).
    \item The $\lambda$-abstraction closes this scope to form the node $g$, which is then wired into \texttt{Rec}.
\end{itemize}

\begin{tikzpicture}[scale=0.8, transform shape,
    node distance=1.5cm and 2cm,
    term/.style={circle, draw=black, very thick, fill=white, minimum size=0.9cm, inner sep=2pt, font=\bfseries},
    ctor/.style={circle, draw=blue!80!black, very thick, fill=blue!10, minimum size=0.9cm, font=\bfseries},
    op/.style={rectangle, draw=red!70!black, fill=red!10, thick, inner sep=5pt, font=\small\ttfamily},
    link/.style={-Latex, thick, draw=gray!80},
    scope/.style={draw=green!50!black, dashed, thick, fill=green!5, rounded corners}
]

% --- 1. Base Case Input ---
\node[term, label=left:Base Case] (base) {0};

% --- 2. The Step Function (Dependency Subgraph) ---
% We construct the lambda body first
\node[term, right=3cm of base, label=below:$y$] (y) {}; % Variable y (previous result)
\node[term, above=0.5cm of y, label=left:$n$] (n) {}; % Variable n (unused)

% Apply Successor twice
\node[ctor, right=1cm of y] (sy) {$S$};
\node[ctor, right=1cm of sy] (ssy) {$S$};

% Wiring inside scope
\draw[link] (y) -- (sy);
\draw[link] (sy) -- (ssy);

% Scope Box
\begin{scope}[on background layer]
    \node[scope, fit=(n) (y) (sy) (ssy), label=below:\textcolor{green!40!black}{\textit{Scope $\mathcal{D}(n, y)$}}] (box) {};
\end{scope}

% The Lambda Abstraction Node
\node[op, right=0.5cm of box] (lam) {$\lambda n.\lambda y$};
\draw[link] (ssy) -- (lam);

% --- 3. The Recursion Operator ---
\node[op, below right=1cm and 1cm of lam, label=right:\textbf{double}] (rec) {Rec$_\mathbb{N}$};

% Wiring Inputs to Rec
\draw[link] (base) to[out=-45, in=180] (rec); % Base Case (0)
\draw[link] (lam) to[out=-90, in=90] (rec);   % Step Case (Function)

% --- Annotations ---
\node[above=0.2cm of ssy, font=\small, color=blue!80!black] {$S(S(y))$};
\node[below=0.5cm of rec, font=\small, align=center] {
    The resulting node \texttt{double}\\
    is a function $\mathbb{N} \to \mathbb{N}$.
};

\end{tikzpicture}

\def\Univ{\textbf{Univ}}

\subsection{Example Construction: Distributivity of $\Pi$ over $\Sigma$}
\label{s:proof_distributivity}

To illustrate the interplay between Formation, Introduction, and Elimination hyperedges within the graph $\St$, we present the proof of the distributivity of dependent functions over dependent pairs given in \S 1.11 of the HoTT book \cite{aczel2013homotopy}.
Logically, this theorem asserts that a function returning a pair is equivalent to a pair of functions.

\textbf{Theorem.} Given a type $A : \Univ$ and families $P, Q : A \to \Univ$, there exists a term of type:
\[
    \left( \Pi_{(x:A)} (P(x) \times Q(x)) \right) \to \left( (\Pi_{(x:A)} P(x)) \times (\Pi_{(x:A)} Q(x)) \right)
\]

\noindent \textbf{Hypergraph Construction Procedure:}

The proof corresponds to the existence of a path of hyperedges transforming the input hypothesis into the target conclusion. We proceed by constructing the necessary nodes in $\St$:

\begin{tikzpicture}[scale=0.8, transform shape,
    % Style definitions based on your Hypergraph formalism
    node distance=1.5cm and 2cm,
    term/.style={circle, draw=black, very thick, fill=white, minimum size=0.8cm, inner sep=2pt, font=\bfseries},
    type/.style={rectangle, draw=blue!80, thick, fill=blue!5, rounded corners, font=\small\itshape, minimum height=0.6cm},
    op/.style={rectangle, draw=red!70!black, fill=red!10, thick, inner sep=4pt, font=\small\ttfamily},
    link/.style={-Latex, thick, draw=gray!80},
    scope/.style={draw=green!50!black, dashed, thick, fill=green!5, rounded corners}
]

% --- 1. The Input Assumption ---
% The term 'f'
\node[term, label=left:$f$] (f) {};
% Its Type (for context)
\node[type, above=0.2cm of f] (type_f) {$\Pi_{(x:A)} (P(x) \times Q(x))$};
\draw[->, blue, dashed] (f) -- (type_f);

% --- 2. The Context/Scope for x:A ---
% We create a scope box later, but let's place the internals first.

% The variable x
\node[term, right=2.5cm of f, label=below:$x$] (x) {};
\node[type, below=0.2cm of x] (type_x) {$A$};
\draw[->, blue, dashed] (x) -- (type_x);

% Operation: App
\node[op, right=1.5cm of f] (app) {App};

% The result of f(x)
\node[term, right=1cm of app, label=above:$f(x)$] (fx) {};

% Wiring App
\draw[link] (f) -- (app);
\draw[link] (x) to[out=135, in=-90] (app); % x flows into App
\draw[link] (app) -- (fx);

% --- 3. Elimination of the Pair (Projections) ---
% Projections operations
\node[op, above right=0.5cm and 1.5cm of fx] (pr1_op) {pr$_1$};
\node[op, below right=0.5cm and 1.5cm of fx] (pr2_op) {pr$_2$};

% Resulting terms
\node[term, right=0.8cm of pr1_op, label=above:$t_1$] (t1) {};
\node[term, right=0.8cm of pr2_op, label=below:$t_2$] (t2) {};

% Wiring Projections
\draw[link] (fx) to[out=45, in=180] (pr1_op);
\draw[link] (fx) to[out=-45, in=180] (pr2_op);
\draw[link] (pr1_op) -- (t1);
\draw[link] (pr2_op) -- (t2);

% --- Draw the Scope Box (The Dependency Subgraph) ---
% This represents the "Lambda" context D(x -> t)
\begin{scope}[on background layer]
    \node[scope, fit=(x) (app) (fx) (pr1_op) (pr2_op) (t1) (t2) (type_x), label=below:\textcolor{green!40!black}{\textit{Dependency Subgraph $\mathcal{D}(x)$}}] (box) {};
\end{scope}

% --- 4. Introduction of Pi (Lambda Abstraction) ---
% The Lambda operations take the whole scope (conceptually) or the specific result relative to x
\node[op, right=1cm of box.north east, anchor=west, yshift=-1cm] (lam1) {$\lambda x$};
\node[op, right=1cm of box.south east, anchor=west, yshift=1cm] (lam2) {$\lambda x$};

% Resulting Functions
\node[term, right=0.8cm of lam1, label=above:$g$] (g) {};
\node[term, right=0.8cm of lam2, label=below:$h$] (h) {};

% Wiring Lambdas
% Graphically, we show the path from the result of the scope to the lambda
\draw[link] (t1) -- (lam1);
\draw[link] (t2) -- (lam2);
% We also conceptually link the variable x to the lambda (binding)
\draw[link, dashed] (x) to[out=0, in=180] (lam1.west);
\draw[link, dashed] (x) to[out=0, in=180] (lam2.west);

\draw[link] (lam1) -- (g);
\draw[link] (lam2) -- (h);

% --- 5. Introduction of Sigma (Pairing) ---
\node[op, right=1.5cm of $(g)!0.5!(h)$] (pair_op) {Pair};

% Final Result
\node[term, right=1cm of pair_op, label=right:Result] (res) {};
\node[type, above=0.2cm of res, align=center] (type_res) {$\Pi P \times \Pi Q$};

% Wiring Final Pair
\draw[link] (g) to[out=-30, in=160] (pair_op);
\draw[link] (h) to[out=30, in=200] (pair_op);
\draw[link] (pair_op) -- (res);
\draw[->, blue, dashed] (res) -- (type_res);

\end{tikzpicture}

\begin{enumerate}
    \item \textbf{Hypothesis (Input Node).}
    We begin with a node $f$ representing the antecedent. This node acts as the root of our transformation:
    \[ f : \Pi_{(x:A)} (P(x) \times Q(x)) \]

    \item \textbf{Context Extension (Entering the Subgraph).}
    To decompose the range of $f$, we must first instantiate its domain. We generate a dependency subgraph $\mathcal{D}(x)$ rooted at the type node $A$. Within this subgraph, $x$ exists as a variable node.

    \item \textbf{Function Elimination (Application).}
    Inside $\mathcal{D}(x)$, we instantiate the \texttt{App} hyperedge. This connects $f$ and $x$ to generate the application node:
    \[ f(x) : P(x) \times Q(x) \]
    Note that this node remains static; no reduction occurs as $x$ is a variable.

    \item \textbf{Pair Elimination (Projection).}
    The node $f(x)$ is of a $\Sigma$-type. We instantiate the pair elimination hyperedges ($\pi_1, \pi_2$) to extract the components. This yields two new nodes within the subgraph:
    \begin{align*}
        t_1 &:= \pi_1(f(x)) : P(x) \\
        t_2 &:= \pi_2(f(x)) : Q(x)
    \end{align*}

    \item \textbf{Function Introduction (Abstraction).}
    We now exit the dependency subgraph $\mathcal{D}(x)$. We apply the $\Pi$-Introduction hyperedge ($\lambda$) to close the scope of $x$ relative to our projection nodes. This generates two function nodes in the outer graph:
    \begin{align*}
        g &:= \lambda x. t_1 \quad \text{(of type } \Pi_{(x:A)} P(x)\text{)} \\
        h &:= \lambda x. t_2 \quad \text{(of type } \Pi_{(x:A)} Q(x)\text{)}
    \end{align*}
    Crucially, the hypergraph formalism treats these $\lambda$-nodes as encapsulations of the entire history within $\mathcal{D}(x)$ that led to $t_1$ and $t_2$.

    \item \textbf{Pair Introduction (Conclusion).}
    Finally, having constructed $g$ and $h$, we instantiate the $\Sigma$-Introduction hyperedge (\texttt{Pair}). This links $g$ and $h$ to form the final result node:
    \[ \text{result} := (g, h) \]
    The typing edge for this result, generated by the pair formation rule, matches the consequent of our theorem.
\end{enumerate}

\noindent \textit{Remark.} This construction demonstrates the ``Conservation of Information'' in the hypergraph. The path from $f$ to $(g, h)$ preserves all dependency data. A computation edge (reduction) exists in the reverse direction: applying projections to $(g,h)$ would structurally reduce back to $f(x)$ via $\beta$-reduction.

\bibliographystyle{plainnat}
\bibliography{bibliography}

\end{document}